\documentclass[journal]{IEEEtran}

\usepackage{xcolor} 
\usepackage{cite} 
\usepackage{amsmath,amssymb,amsfonts,bbm} 
\usepackage{algorithmic}
\usepackage{tabularx} 
\usepackage{amsmath, bm} 
\usepackage{amsmath, autobreak} 
\usepackage[ruled, linesnumbered]{algorithm2e} 
\usepackage{subfigure} 
\usepackage{soul} 
\usepackage{balance} 
\usepackage{color} 
\usepackage{float} 
\usepackage{autobreak} 
\usepackage{graphicx} 
\usepackage{textcomp} 
\usepackage{xcolor} 
\usepackage{booktabs} 
\usepackage{multirow} 
\usepackage{bbm} 
\usepackage{dsfont} 
\usepackage{footnote} 
\usepackage[]{mathtools} 
\usepackage{lipsum} 
\makesavenoteenv{algorithm} 
\usepackage{diagbox}
\usepackage{lineno}
\usepackage{tikz}
\usepackage{makecell}
\usepackage{amssymb}
\usepackage{mathrsfs}
\usepackage{calligra}
\usepackage[numbers,sort&compress]{natbib}

\usepackage[colorlinks, linkcolor=black, anchorcolor=black, citecolor=black]{hyperref}

\begin{document}

\title{Secure Deep Reinforcement Learning for Dynamic Resource Allocation in Wireless MEC Networks}

\author{    
    Xin~Hao, 
    Phee~Lep~Yeoh,~\IEEEmembership{Senior~Member,~IEEE,} 
    Changyang~She,~\IEEEmembership{Senior~Member,~IEEE,} 
    Branka~Vucetic,~\IEEEmembership{Life~Fellow,~IEEE,} 
    and~Yonghui~Li,~\IEEEmembership{Fellow,~IEEE}
\thanks{The work of P. L. Yeoh was supported in part by The University of Sydney Robinson Fellowship. The work of C. She was supported in part by DECRA under Grant DE210100415. The work of Branka Vucetic was supported in part by the ARC Laureate Fellowship grant number FL160100032. The work of Yonghui Li was supported by ARC under Grant DP190101988 and DP210103410. \emph{(Corresponding author: Changyang She.)}}      
\thanks{X. Hao, C. She, B. Vucetic, and Y. Li are with the School of Electrical and Information Engineering, University of Sydney, NSW 2006, Australia (e-mail: xin.hao@sydney.edu.au; shechangyang@gmail.com; branka.vucetic@sydney.edu.au; yonghui.li@sydney.edu.au).

P. L. Yeoh is with the School of Science, Technology and Engineering, University of the Sunshine Coast, QLD 4502, Australia (e-mail: pyeoh@usc.edu.au).
}
}

\markboth{Transactions on Communications}%
{Shell \MakeLowercase{\textit{et al.}}: Bare Demo of IEEEtran.cls for IEEE Journals}

\maketitle

\begin{abstract}
This paper proposes a blockchain-secured deep reinforcement learning (BC-DRL) optimization framework for {data management and} resource allocation in decentralized {wireless mobile edge computing (MEC)} networks. In our framework, {we design a low-latency reputation-based proof-of-stake (RPoS) consensus protocol to select highly reliable blockchain-enabled BSs to securely store MEC user requests and prevent data tampering attacks.} {We formulate the MEC resource allocation optimization as a constrained Markov decision process that balances minimum processing latency and denial-of-service (DoS) probability}. {We use the MEC aggregated features as the DRL input to significantly reduce the high-dimensionality input of the remaining service processing time for individual MEC requests. Our designed constrained DRL effectively attains the optimal resource allocations that are adapted to the dynamic DoS requirements. We provide extensive simulation results and analysis to} validate that our BC-DRL framework achieves higher security, reliability, and resource utilization efficiency than benchmark blockchain consensus protocols and {MEC} resource allocation algorithms.
\end{abstract}

\IEEEpeerreviewmaketitle

\section{Introduction}
\IEEEPARstart{S}{{ecurity}} {is a major concern in mobile edge computing (MEC) service provisioning given the prevalence of data tampering attacks leading to disruptive denial-of-service (DoS) in decentralized wireless networks~\cite{reviewer2_security_analysis, RichardYu_LongSurvey_BC_ML}}. Blockchain-based data management has been recently considered to prevent data tampering attacks and ensure the integrity of MEC user requests in wireless networks~\cite{TWC_blockchain_DRL}. In {blockchain-secured MEC} networks, {the base stations (BSs)} are also blockchain nodes which store critical data at all nodes in the blockchain network according to a given consensus protocol. The most well-known blockchain consensus protocol is proof-of-work (PoW)~\cite{Bitcoin_Nakamoto}, which is not suitable for wireless MEC networks with limited {BS} computation resources and strict latency requirements~\cite{TCOM_consensus_delay}. In~\cite{TCOM_PoD, BC_high_resource_consume}, it was shown that the PoW consensus incurred lengthy delays due to high computations for block validation and requiring all nodes to compete in the block generation. 

{Significant research efforts have focused on reducing} the high computation overhead and processing latency for {blockchain consensus~\cite{PBFT_ref, survey_PoS_no_random_node_select, rep_trustManage_highIndex, TCOM_BC_DoS_MEC, Xin_DBC_IoTJ, TVT_Rep_Contract, Niyato_repVoting_shard}. Among them, a popular approach is the} practical Byzantine fault tolerance (PBFT) consensus {protocol, which reduces the block generation time by selecting} one blockchain node to generate {a} new block~\cite{PBFT_ref}. PBFT consensus still has high computation and latency for block validation, because all nodes need to validate the generated block. Proof-of-stake (PoS) consensus {is another protocol that can reduce computation and latency overheads by} selecting a trusted subset of blockchain nodes to validate the generated block~\cite{survey_PoS_no_random_node_select}. However, PoS consensus is vulnerable to attacks since the highest stake miner node that is selected for block generation can be easily targeted by attackers~\cite{rep_trustManage_highIndex}. Clearly, there is a pressing need to {design a novel blockchain consensus that can} significantly reduce the high computation overheads {without jeopardizing the security level.}

{Apart from security, a further challenge in dynamic MEC networks is to efficiently allocate the computation resources in each time slot~\cite{TCOM_invited_model_learn}, since} allocating more computation resources in the current time slot results in a smaller processing latency for these users, but a potentially higher DoS probability due to insufficient resources for future users. This {fundamental} trade-off between processing latency and DoS probability in MEC networks can be managed by formulating the optimal resource allocation as a sequential decision-making problem~\cite{Survey_ML_MEC, TCOM_DRL_diverse_QoS}. {Dynamic programming is a traditional model-based approach used for resource allocation in sequential decision-making problems~\cite{book_RL_DP}. However, it is challenging to apply in large-scale problems due to the exponential growth of state and action spaces~\cite{Chapter_Curse_dimension}. To overcome this challenge, deep reinforcement learning (DRL) algorithms have been employed~\cite{TCOM_DRL_UAV, letter_DDPG_MEC}, and constrained DRL is an effective  solution to address the explicit requirements on constraints by reformulating the original optimization as a constrained Markov decision process (MDP)~\cite{SCY_tutorial_urllc}. Recently, there is an urgent need to consider security constraints in DRL with the emergence of blockchain-secured MEC networks~\cite{reviewer2_secure_DRL}.}

\subsection{Related Works}
{Most MEC blockchain research has focused on enhancing the security of resource-efficient PoS-based consensus protocols. The authors of~\cite{rep_trustManage_highIndex} proposed a secure network management scheme by designing a PoS-based consensus with random node selection in each block generation. To prevent data tampering attacks by the edge node, the authors of}~\cite{TCOM_BC_DoS_MEC} employed a joint PoW and PoS consensus protocol storing the reputations of edge devices in the blockchain. In~\cite{Xin_DBC_IoTJ, TVT_Rep_Contract, Niyato_repVoting_shard}, it was shown that carefully evaluating the reputation values to select highly reliable blockchain nodes for blockchain management can effectively reduce the computation overhead and resist potential blockchain attacks. Since reputation is a long-term evaluation metric, the authors in~\cite{Xin_DBC_IoTJ} {identified} malicious users by evaluating their reputations based on current user data and historical reputations stored in a blockchain. To ensure secure miner selection, the authors in~\cite{TVT_Rep_Contract} used a multiweight model considering past interactions {with} other vehicles to evaluate trusted reputations of blockchain nodes. In~\cite{Niyato_repVoting_shard}, the reputations of MEC BSs acting as blockchain nodes are evaluated based on feedback from both their users and other MEC BSs.

Some recent research efforts have focused on optimizing resource allocation in blockchain-secured MEC networks~\cite{TCOM_game_PoW_MEC, Blockchain_no_AI_TWC, RichardYu_MDP_SNRstate}. In~\cite{Blockchain_no_AI_TWC}, {an iterative optimization approach was used} to minimize the weighted sum of MEC energy consumption and blockchain latency in an MEC system. In~\cite{TCOM_game_PoW_MEC}, the authors analyzed a PoW-based consensus protocol {and used a game-theoretic optimization framework to solve} the target non-cooperative resource allocation. {To improve the efficiency of resource allocation, more recent research has proposed to apply secure DRL in MEC service provisioning. In~\cite{RichardYu_MDP_SNRstate}, the authors {used an unconstrained DRL approach} to maximize the weighted sum of blockchain throughput and the reciprocal of MEC delay.} However, unconstrained DRL may encounter difficulties in explicitly satisfying dynamic constraints, which can be addressed by transforming the problem into the dual domain to optimize the weights between the objective and constraints~\cite{Lagrangian_PrimalDual}. In~\cite{SCY_VR_GC_2021}, constrained DRL was applied in a virtual reality network, where the weight between the video loss ratio and processing latency is optimized. {To further improve the training efficiency of DRL, researchers have explored methods to improve training efficiency, such as choosing low dimension features to reduce the complexity of the optimization problem~\cite{SCY_digital_twin}, and applying transfer learning of pre-trained parameters when new MEC devices join the network~\cite{TransferDRL_MEC}. How to improve security and training efficiency for DRL in dynamic MEC networks still remains an open problem for further investigation.}

\subsection{Contributions}
In this paper, we propose a blockchain-secured DRL (BC-DRL) optimization framework {for} decentralized {dynamic} wireless MEC network{s}. The main contributions are summarized as follows:
\begin{itemize}
\item {We propose a reputation-based proof-of-stake (RPoS) blockchain consensus protocol that significantly reduces block generation and validation time whilst maintaining a high level of security by randomly selecting the miner BS node from a subset of BSs with high reputations. Attacks from malicious BSs are prevented by isolating the BSs with low reputations, whilst attacks from malicious users are resisted by using Bayesian inference to evaluate all user feedback. The secure storage of users' MEC requests in blockchain-enabled BSs further mitigates tampering attacks targeting MEC service provisioning.} 
\item {We solve the dynamic resource allocation by formulating it as an MDP that minimizes processing latency while satisfying the constraint on DoS probability. This formulation optimizes the allocation of computation resources for both blockchain consensus and MEC service provisioning. We provide mathematical proofs demonstrating the equivalence of the original problem and the reformulated MDP problem. In addition, we establish that the reformulated MDP satisfies the Markovian property.}
\item {We design a constrained DRL algorithm that can accommodate dynamic requirements on DoS probabilities. To improve the training efficiency, we propose an aggregating mechanism to reduce the dimension of the features as the remaining processing time of all the requests. Transfer learning is utilized to update pre-trained parameters when the requirement on DoS probability changes, and empirical convergence analysis is provided.} 
\end{itemize}
Numerical examples are provided to demonstrate the high-secur{ity}, high-reliability, resource-saving, and low-latency advantages of our BC-DRL solution. {We present detailed analysis and p}erformance comparisons with {existing PoS} consensus protocol and DRL-based resource allocation algorithms, giving insights for implementing future secure blockchain and DRL-empowered dynamic resource allocations.

\begin{figure}[t] 
\centering 
\centerline{\includegraphics[scale=0.85]{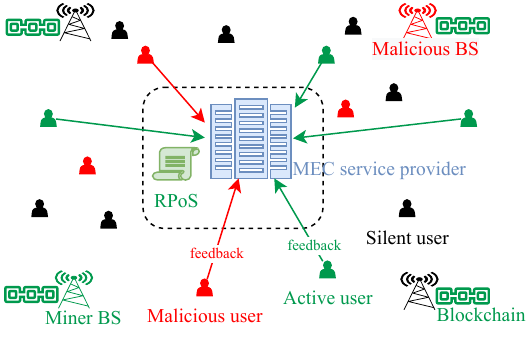}} 
\caption{Our RPoS blockchain consensus selects trusted BSs for MEC service provisioning and blockchain management using feedback from all users to prevent BS denial-of-service (DoS) attacks from both malicious BSs and users.}
\label{fig: CommunicationModel}
\end{figure}
\section{System Model~\label{sec_systemModel}}
In our model, an MEC service provider employs $N_\text{B}$ BSs with overlapping coverage to satisfy the dynamic time-varying requests for computation resources from multiple users with DoS probability constraints for the BSs. In each time slot, we assume each user either stays silent or sends a request with a random workload to the service provider. The workload is defined as the required CPU cycles to complete a user's task. Thus, the number of user requests and the CPU cycles required can be modeled as two independent arrival processes. {In each time slot, one BS is selected by the MEC service provider to process the received MEC requests from the users.} Depending on the available resources, the BS applies the DRL algorithm to allocate or deny resources to these requests in the current time slot. The RPoS consensus is used to securely store the user requests and select BSs to serve the users.

\subsection{Attack Model~\label{section_attack_model}}
Fig.~\ref{fig: CommunicationModel} depicts the considered decentralized wireless MEC network in a specific time slot. The green and black BSs are trusted BSs that participate in blockchain management and MEC service provisioning in the current time slot, while the red BSs are untrusted. The green and black users are non-malicious users sending truthful feedback of the BS service provisioning, while the red users are malicious users. We assume that the majority of BSs and users are non-malicious, {and user feedback is used to evaluate the BS reputation and DoS probability.} 
Our user feedback-based approach helps the service provider to independently identify malicious BSs and is different from previous wireless blockchain consensus designs, where the BS is responsible for evaluating the reputations of the blockchain nodes. 

The attack models for the miner BS attacks and malicious user attacks are detailed as follows.

\begin{table}[t] 
\renewcommand\arraystretch{1} 
\caption{Possible Feedback From Individual Users} 
\centering 
\begin{tabular}{l | c c c c c} 
\toprule 
\toprule 
\textbf{Case}                                                                                               &1       &2       &3         &4         &5\\
\toprule 
\textbf{Active user with request}                                                                           &Yes     &Yes     &Yes       &Yes       &No\\
\hline
\textbf{Service provided}                                                                                   &Yes     &No      &No        &Yes       &Yes/No\\
\hline
\textbf{Feedback from user}                                                                                 &1       &0       &1         &0         &0/1 \\
\hline
\textbf{The user feedback is malicious}                                                                     &No      &No      &Yes       &Yes       &Yes \\
\bottomrule 
\bottomrule 
\end{tabular} 
\label{tab: user_feedback} 
\end{table}

\subsubsection{Miner BS Attacks}
We consider an attacker aims to launch a blockchain 
attack on the miner BS node by monopolizing all the incoming and outgoing connections from the miner BS to the surrounding BSs~\cite{EclipseAttack}. In existing PoS protocols, an attacker can identify the miner BS once it has successfully deciphered the stake evaluation mechanism since the highest stake miner is always selected~\cite{rep_trustManage_highIndex}. Our proposed RPoS mitigates this attack by randomly selecting the miner BS from a subset of high reputation BSs to provide services to the requesting users. 

\subsubsection{Malicious User Feedback Attacks}\label{section_malicious_user_feedback_attack}
We consider that each user sends feedback indicating whether their requests in the current time slot were served or not.  The green users send feedback indicating whether their requests were served in the previous time slot, whilst the black users had no requests and do not send any feedback. The red users are malicious users with or without requests, which are aiming to disrupt the BS reputation evaluation by sending untruthful feedback.  Table~\ref{tab: user_feedback} summarizes all possible user feedback in different cases.

\begin{figure}[t]
\centering 
\centerline{\includegraphics[scale=0.85]{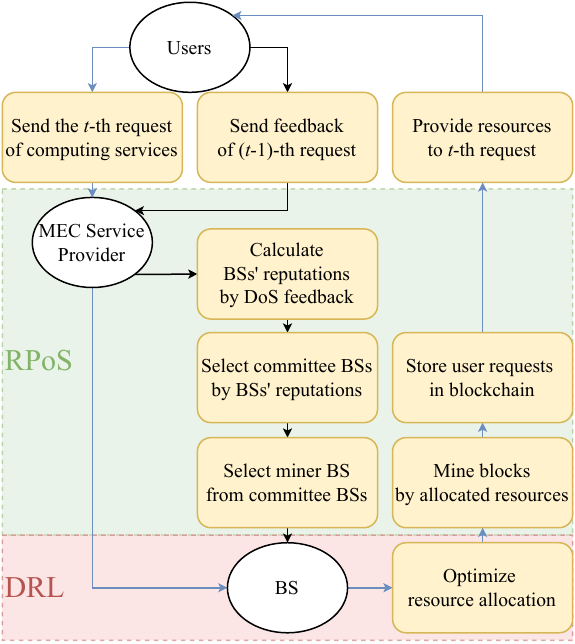}} 
\caption{Blockchain-secured deep reinforcement learning (BC-DRL) framework for efficient and secure resource allocation.} 
\label{fig_RPoS_management}
\end{figure}

\subsection{BC-DRL Solution}
Fig.~\ref{fig_RPoS_management} shows the decentralized architecture of our BC-DRL solution, which includes three entities: 1) Users, 2) MEC service provider managing the RPoS consensus, and 3) BSs implementing the DRL-based resource allocation algorithm. Their respective actions are detailed as follows.

\subsubsection{Users}
In each time-slot, a random number of users send computing service requests to the MEC service provider. The users may also send feedback to the service provider indicating if their requests in the previous time slot were satisfied.

\subsubsection{MEC Service Provider}
First, the feedback received from the users are utilized to evaluate the BSs' reputations by Bayesian inference. Then, to improve the resource utilization efficiency without jeopardizing the system security level, trusted committee BSs with high reputations are selected to manage the blockchain network. Furthermore, an optimal BS is selected as the miner BS providing computing services to the users in each time slot. Lastly, all the user requests are packaged in a new block and stored in the blockchain to prevent data tampering attacks from changing the user requests and enhance the reliability of the DRL-based resource allocation.

\subsubsection{BS}
Implements the DRL algorithm optimizing computation resource allocation for the selected miner BS to mine the new block and serve user requests. 

\section{RPoS-Based Blockchain Management~\label{sec_architecture}}
In this section, we introduce the BS reputation evaluation to select trusted BSs. Next, we outline the RPoS consensus protocol and derive the CPU cycles for block generation and commitment which is used to evaluate the blockchain processing latency. {Lastly, we analyze the tampering attack-resistant ability and discuss the trade-off between security and resource consumption.}

\subsection{BS Reputation Evaluation for Blockchain Consensus}~\label{section_rep_evaluate}
The BS reputation evaluation of our RPoS consensus is based on Bayesian inference of the aggregated user feedback. We denote $e_{i}(t)$ as the event that the requests in the $t$-th time slot are served by the $i$-th BS, and $\Pr\{e_{i}(t)\}$ as the prior probability of event $e_{i}(t)$. Similarly, we denote $\overline{e}_{i}(t)$ as the complementary event of $e_i(t)$ that requests are denied by the $i$-th BS. Hence, we can obtain that $\Pr\{\overline{e}_{i}(t)\}=1-\Pr\{e_{i}(t)\}$.

\subsubsection{User Feedback Mechanism}
We define $\mathcal{K}(t)$ as the set of users sending feedback in the $t$-th time slot, and denote $d_{i,{k}}(t)$ as the feedback from the ${k}$-th user. Specifically, $d_{i,{k}}(t)=0$ indicates that the $k$-th user is served by the $i$-th BS, and $d_{i,k}(t)=1$ indicates that the service is denied.

\begin{figure}[t] 
\centering 
\centerline{\includegraphics[scale=0.45]{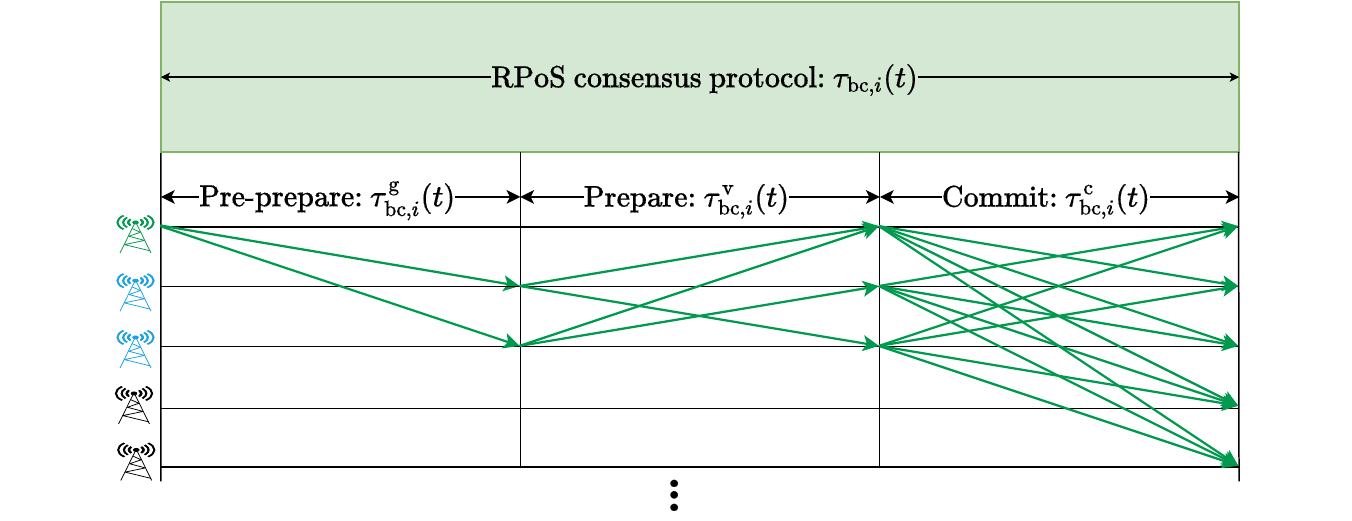}} 
\caption{RPoS consensus for generating, validating, and committing a new block where the green BS is the miner BS, the blue BSs are the validator BSs, and the black BSs are the remaining BSs in the network.}
\label{fig: consensus} 
\end{figure}

\subsubsection{Bayesian-Based DoS Inference Evaluation}

Since malicious users may send feedback without previously making any requests, it is necessary to evaluate the probability that the BS has successfully served the user requests, which is also known as the DoS inference. The received feedback from the users in the $t$-th time slot is denoted by $\mathcal{D}_{i,\mathcal{K}}(t)=\{d_{i,1}(t), d_{i,2}(t),\cdots, d_{i,K}(t)\}$. Given the observation of $\mathcal{D}_{i,\mathcal{K}}(t)$, we obtain the DoS inference as the probability that the user requests are served by the $i$-th BS in the $t$-th time slot by Bayesian inference as~\cite{rep_trustManage_highIndex, Bayesian_inference_draft}
\begin{equation}
I_{i}(t)\triangleq 
{\Pr}\{e_{i}(t)|\mathcal{D}_{i,\mathcal{K}}(t)\}  
=\frac{\Pr\{e_{i}(t)\} \Pr\{\mathcal{D}_{i,\mathcal{K}}(t)|e_{i}(t)\}} 
{\Pr\{\mathcal{D}_{i,\mathcal{K}}(t)\}},
\label{eq: Bayesian_current_information}
\end{equation} 
where $\Pr\{\mathcal{D}_{i,\mathcal{K}}(t)\}=\Pr\{e_{i}(t)\} \Pr\{\mathcal{D}_{i,\mathcal{K}}(t)|e_{i}(t)\}+\Pr\{\overline{e}_{i}(t)\} \Pr\{\mathcal{D}_{i,\mathcal{K}}(t)|\overline{e}_{i}(t)\}$ is the prior probability of $\mathcal{D}_{i,\mathcal{K}}(t)$. Since different users generate feedback independently, we have $\Pr\{\mathcal{D}_{i,\mathcal{K}}(t)|e_{i}(t)\}=\prod\nolimits_{k \in \mathcal{K}(t)}   \Pr\{d_{i,k}(t)|e_{i}(t)\}$ and $\Pr\{\mathcal{D}_{i,\mathcal{K}}(t)|\overline{e}_{i}(t)\}=\prod\nolimits_{k \in \mathcal{K}(t)} \Pr\{d_{i,k}(t)|\overline{e}_{i}(t)\}$. We denote $\Pr\{d_{i,k}(t)|e_{i}(t)\}$ and $\Pr\{d_{i,k}(t)|\overline{e}_{i}(t)\}$ as conditional probabilities that the requests are served under the conditions of $e_i(t)$ and $\overline{e}_{i}(t)$, respectively.

\subsubsection{BS Reputation Evaluation}
The reputation in the $t$-th time slot is updated according to
\begin{align} 
\xi_{i}(t)= 
\begin{cases} 
\xi_{i}(t-1),                 & \text{if }  \mathcal{D}_{i,\mathcal{K}}(t) =\varnothing \\ 
\xi_{i}^\text{a}(t),    & \text{if } \mathcal{D}_{i,\mathcal{K}}(t) \neq\varnothing \\ 
\end{cases} 
\label{eq_Rep_update} 
\end{align}
where $\xi_{i}^\text{a}(t)$ represents the updated reputation when user feedback exists. We note that eq.~\eqref{eq_Rep_update} indicates the $i$-th BS in the $t$-th time slot keeps the same value as that in the $(t-1)$-th time slot if there is no feedback received in the $t$-th time slot, otherwise evolves to an updated value,  $\xi_{i}^a(t)$, which is a weighted sum of the current Bayesian inference and the historical reputations, i.e.,
\begin{equation} 
\begin{split} 
\xi_{i}^\text{a}(t)= 
\vartheta_\text{I} I_{i}(t)  
+(1-{\vartheta_\text{I}}) {\xi}_{i}^\text{h}(t),
\label{eq: reputation_update} 
\end{split} 
\end{equation} 
where $\vartheta_\text{I}$ is {defined as the weight coefficient of reputation inference indicating the weight of the Bayesian inference}, and ${\xi}_{i}^\text{h}(t)={1}/{{\tau}_{\xi}}\sum\nolimits_{t'=t-1}^{t-{\tau}_{\xi}} \beta_{\xi}(t-t')\xi_i\left(t'\right)$ is the expected influence of historical reputations in the past ${\tau}_{\xi}$ time slots~\cite{Xin_WFIoT}, where $\beta_{\xi}(t-t')$ is the discount factor of the historical reputations in the $(t-t')$-th time slot (e.g. $\beta_{\xi}(t) = e^{- t}$~\cite{TrustChain_ICBC}, $\beta_{\xi}(t) = (1/2)^{t}$, $\beta_{\xi}(t) = t^{-1}$).

\subsection{Proposed RPoS Consensus Protocol}

Fig.~\ref{fig: consensus} shows the three main steps of the RPoS consensus protocol for generating, validating, and committing a new block to the blockchain. Specifically, in the pre-prepare step, the MEC service provider assigns one BS from the committee as the miner BS to generate a new block by computing a unique hash signature based on the data prepared for packaging in the block. Next, in the prepare step, the newly generated block is validated by all the other BSs in the committee, known as validator BSs. To do so, each validator BS computes their signatures for comparison with the hash signatures generated by all the other committee BSs. Lastly, in the commit step, the new block is stored in all the BSs in the network. We define $\mathcal{N}_\text{M}(t)$ and $\mathcal{N}_\text{V}(t)$ as the sets of the committee miner BS and validator BSs in the $t$-th time slot. 

Next, we detail the proposed RPoS consensus protocol, which involves the trusted committee BS and miner BS selection process using our BS reputations evaluated in {eq.}~\eqref{eq: reputation_update}. 

\subsubsection{Committee BSs Selection}
In our BC-DRL framework, we select a subset of BSs with reputations higher than a threshold to participate in the proposed RPoS consensus protocol. We denote ${\xi}_{\eta}(t)$ as the threshold reputation value, which can be evaluated as
\begin{equation} 
\xi_{\eta}(t)=\eta \bar{\xi}(t)=\dfrac{\eta}{N_\text{B}}  {\sum\nolimits_{i \in \mathcal{N}_\text{B}} {\xi_{i}(t)}}, 
\label{eq: reputation_threshold} 
\end{equation} 
where $\overline{\xi}(t)$ is defined as the average reputation of the overall BSs in the network in the $t$-th time slot, and $\eta$ is {defined as the weight coefficient of the reputation threshold}. A larger value of $\eta$ corresponds to a higher reputation requirement of the committee BSs. Based on~\eqref{eq: reputation_threshold}, the number of BSs elected to the blockchain committee in the $t$-th time slot be calculated by
\begin{equation} 
N_\text{C}(t) = \sum\nolimits_{i \in \mathcal{N}_\text{B}}   \mathds{1}\{{\xi_i(t)} \geq  {\xi}_{\eta}(t)\},  
\label{eq: num_committee_BS}
\end{equation} 
where $\mathds{1}\{\cdot\}$ is the indicator function, which equals to one if ${\xi_i(t)} \geq {\xi_{\eta}}(t)$, and equals to zero otherwise. The subset of BSs with reputations higher than $\xi_{\eta}(t)$ is referred to as the blockchain committee in BC-DRL.

\subsubsection{Miner BS Selection}
In RPoS, we consider that the miner BS is randomly selected from the subset of trusted committee BSs in each time slot to mitigate against miner BS attacks. As such, the probability of an attacker identifying the miner BS is
\begin{equation}
p_\mathrm{M}(t) 
\triangleq \Pr\{\mathcal{N}_\text{A}(t)=\mathcal{N}_\text{M}(t)\}
=\frac{N_\text{M}(t)}{\mathbb{E}[N_\text{C}(t)]},
\label{eq_RPoS_attack_probability}
\end{equation}
where $\mathcal{N}_\text{A}(t)$ indicates the set of BSs under attack, and $\mathbb{E}[\cdot]$ denotes the expectation operation. We observe that the probability of successful attacks on the miner BS decreases with increasing size of the blockchain mining committee, $N_\mathrm{C}(t)$.

\subsection{CPU Cycles Required by Miner BS and Blockchain Processing Latency}~\label{sec_CPU_cycle_blockchain}
We can observe from Fig.~\ref{fig: consensus}, the miner BS performs computations in the pre-prepare and commit steps. If the $i$-th BS is assigned as a miner in the $t$-th time slot, the required CPU cycles for block generation and validation is given by
\begin{equation} 
\begin{split} 
f_{\text{bc},i}(t)  
&=f_{\text{bc},i}^\text{g}(t)+f_{\text{bc},i}^\text{c}(t) \\
&=\mathds{1}\{i \in \mathcal{N}_\text{M}(t)\} {\kappa}_\text{bc} {\ell_\text{b}(t)} ( 1 +  N_\text{V}(t) )
,
\text{ (CPU cycles)} 
\label{eq_computing_resource_bc_mine} 
\end{split} 
\end{equation} 
where $f_{\text{bc},i}^\text{g}(t)$ and $f_{\text{bc},i}^\text{c}(t)$ are the CPU cycles required by the pre-prepare and commit steps, respectively. The miner BS needs to calculate the hash value for its unique signature to be added to the new block in the pre-prepare step, and also calculates all the signatures of committee validator BSs in the block in the commit step.

We note that the CPU cycles required by the miner BS {are} proportional to the real-time block size, which is defined by
\begin{equation} 
\ell_\text{b}(t)=\ell_\text{h} + \ell_\text{body}(t),  \quad  \text{(bytes)} 
\label{eq: block_size} 
\end{equation} 
where $\ell_\text{h}$ is the size of the block header, which has a fixed size. The $t$-th block header stores the hash value of the $(t-1)$-th block. The blockchain structure ensures the data in each block is resistant to tampering attacks. The size of the block body is given by $\ell_\text{body}(t)=\ell_\text{c} {\lambda} (t)$ (bytes), which is proportional to the number of user requests in the $t$-th time slot. The coefficient $\ell_{\rm c}$ is a constant, and $\lambda(t)$ is the number of requests generated by all the users in the $t$-th time slot.

Based on eqs.~\eqref{eq_computing_resource_bc_mine} and~\eqref{eq: block_size}, the blockchain processing latency for the RPoS consensus is given by
\begin{equation} 
\tau_{\text{bc},i}(t)=\tau_{\text{bc},i}^\text{g}(t)+\tau_{\text{bc},i}^\text{v}(t)+\tau_{\text{bc},i}^\text{c}(t), \quad\text{(slots)} 
\label{eq: Latency_RPoS} 
\end{equation} 
where $\tau_{\text{bc},i}^\text{g}(t)$, $\tau_{\text{bc},i}^\text{v}(t)$ and $\tau_{\text{bc},i}^\text{c}(t)$ are the processing latency introduced by the pre-prepare, prepare, and commit steps, respectively. 
The detailed derivation of $\tau_{\text{bc},i}(t)$ in \eqref{eq: Latency_RPoS} which includes the block computing and wireless transmission time amongst all the BSs is given in Appendix~\ref{Processing_latency_blockchain}.

\subsection{{Security Analysis}}
{The ability of BC-DRL to resist tampering attacks is a crucial aspect of secure MEC service provisioning. To analyze the security performance, we evaluate the minimum time required to tamper the blockchain, which is defined as~\cite{Xin_DBC_IoTJ}
\begin{equation}
    \tau_\mathrm{tam}(t) =\frac{N_\text{B}}{2} {\mathbb{E}_i[\tau_{\text{bc},i}(t)]}, \quad i \in N_\mathrm{C}(t),
\end{equation}
where the expectation is taken over the selected committee members in the $t$-th time slot. This equation reveals that the MEC service provisioning with blockchain has a very high tampering attack-resistant ability, as an attacker would need to compromise at least half of the BSs in the network to succeed in tampering with the data. This robust security feature ensures the integrity and reliability of BC-DRL for MEC services.}

{Based on our analysis, integrating blockchain into MEC introduces additional latency and computation resources. However, these trade-offs are justified by the benefits, such as tamper attack resistance and steady service provisioning. For example, the RPoS consensus protocol is resource-efficient, distinguishing it from existing blockchain consensus protocols. The dynamic DoS probability constraint ensures security and stability, effectively mitigating potential attacks.}

\begin{figure}[t] 
\centering 
\centerline{\includegraphics[scale=0.45]{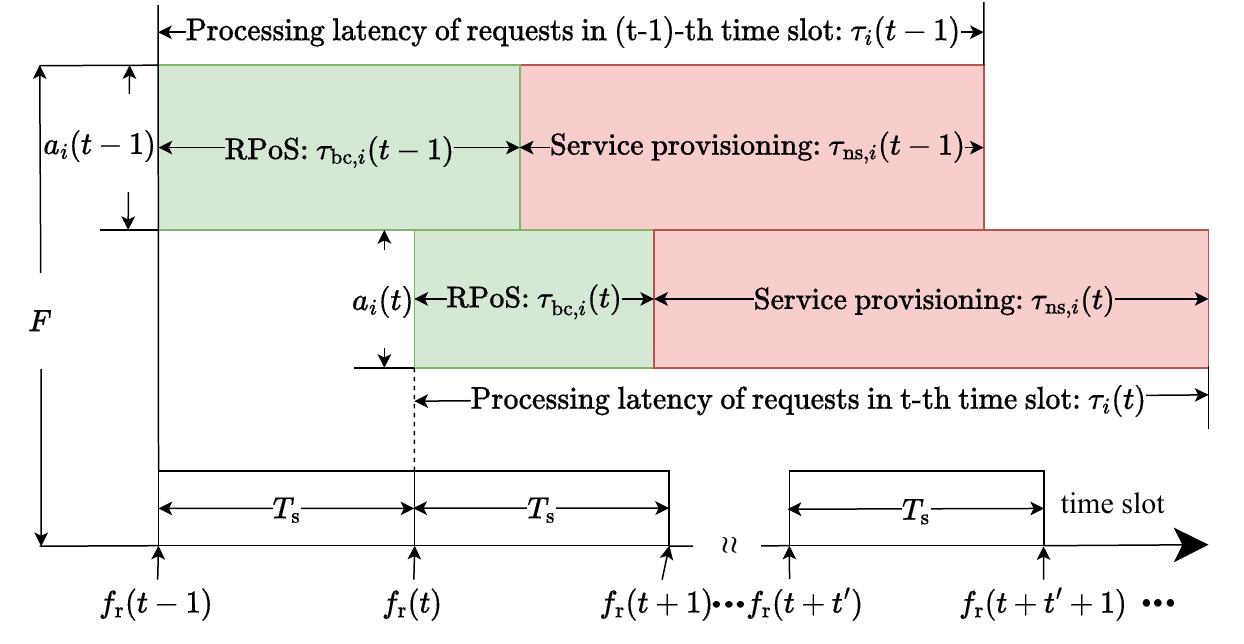}} 
\caption{Example of allocated service rates, $a_{i}(t)$, and the overall processing latency, $\tau_{i}(t) = \tau_{\mathrm{bc},i}(t) + \tau_{\mathrm{sp},i}(t)$, where $F$ is the total computation capacity of each BS, $T_s$ is the duration of one time slot, and $f_\mathrm{r}(t)$ is the total number of requested CPU cycles in the $t$-th time slot.}
\label{fig: time_relationship} 
\end{figure}
\section{Constrained DRL Resource Optimization~\label{section_DRL_algrithm}}
In this section, we present our MEC computation resource allocation optimization to minimize the overall processing latency for the blockchain and MEC service provision{ing} subject to constraints on the BS DoS probability. We formulate the optimization problem as a constrained MDP, which is solved using a constrained DRL algorithm. To improve the training efficiency, we reduce the dimension of the neural network's inputs and apply transfer learning to handle changes in the constraints on DoS probability. {Lastly, we give the complexity analysis of our proposed constrained DRL algorithm.}

\subsection{BC-DRL Optimization}
We aim to minimize the overall processing latency for our BC-DRL framework whilst guaranteeing a given BS DoS probability constraint. The optimization problem {is} formulated as
\begin{align} 
\begin{split} 
&\min_{a_{i}(t)}\quad \mathbb{E}[\tau_{i}(t)]\\
&\begin{array}{r@{\quad}l@{}l@{\quad}l} 
\text{s.t.}\quad & \mathbb{E}[{c}_{i}(t)] {\leq \epsilon_{\max}},
\label{eq: formulated_optimization} 
\end{array} 
\end{split} 
\end{align}
where $\tau_{i}(t)$ is expressed in~\eqref{eq_processing_latency} indicating the overall processing latency for the $i$-th BS, and ${c}_{i}(t)$ expressed in~\eqref{eq_DoS_probability} is the BS DoS probability for the $i$-th BS indicating that the assigned miner BS does not allocate any resources in the $t$-th time slot. 

In Fig.~\ref{fig: time_relationship}, we see that the overall processing latency depends on the number of requested CPU cycles $f_\text{r}(t)$ in each time slot with duration $T_\text{s}$ and the allocated service rate $a_{i}(t)$ (CPU cycles/slot). Given the fixed maximum computation capacity of a BS, there is a trade-off between $\tau_{i}(t)$ and ${c}_{i}(t)$. Based on the time-varying $f_\text{r}(t)$, we optimize the allocated service rate, $a_i(t)$, to minimize the average processing latency subject to the average DoS probability constraint.

\subsubsection{Overall Processing Latency}
We consider the service arrival process of each user request follows an independent and identically distributed (i.i.d) Bernoulli process. As such, the total number of user requests in each time slot follows a Poisson distribution, and the total number of CPU cycles required to satisfy all user requests in the $t$-th time slot can be evaluated as
\begin{equation} 
\begin{split} 
f_\text{r}(t)={\kappa}_\text{sp}\cdot\lambda(t)\sum_{u=1}^{  {\lambda}(t)} {  \ell_{u}(t)}, \quad \text{(CPU cycles)} 
\label{eq: resource_ue_req} 
\end{split} 
\end{equation} 
where $\kappa_\text{sp}$ (CPU cycles/byte) is the coefficient of CPU cycles required for service {provisioning}, ${\lambda}(t)$ is the Poisson distribution parameter representing the average number of user requests, and $\ell_{u}(t)$ (bytes/request) is the package size of the $u$-th requesting user.
Since the miner BS is designed to provide services to the users, the total CPU cycles required for the service {provision} BS can be expressed as
\begin{equation} 
f_{\text{sp},i}(t)=\mathds{1}\{i \in \mathcal{N}_\text{M}(t)\} \cdot f_\text{r}(t). \quad \text{(CPU cycles)} 
\label{eq: CPU_cycles_for_network_slicing} 
\end{equation}

The overall processing latency {corresponds to the summation latency of the processing the blockchain and the MEC service by the miner BS indexed by $i$, and is} given by
\begin{equation} 
\tau_{i}(t) = \tau_{\text{bc},i}(t) + \tau_{\text{sp},i}(t),   \quad \text{(slots)} 
\label{eq_processing_latency} 
\end{equation}
where $\tau_{\text{bc},i}(t)$ is blockchain processing latency given in eq.~\eqref{eq: Latency_RPoS} and $\tau_{\text{sp},i}(t)={f_{\text{sp},i}(t)}/{a_{i}(t)}$ is the processing latency of the MEC service {provision}, where $a_{i}(t)$ (CPU cycles/slot) is the service rates allocated in the $t$-th time slot by the {miner BS, referred to the} $i$-th BS.

\subsubsection{BS DoS Probability}
The instantaneous BS DoS indicator of the $i$-th BS in the $t$-th time slot is defined as
\begin{equation}
c_i(t)= \mathds{1}\{a_i(t)=0\},
\label{eq_DoS_probability}
\end{equation}
where $\mathds{1}\{\cdot\}$ is the indicator function, which equals one when no resources are allocated in the $t$-th time slot by the $i$-th BS (i.e.,  $a_{i}(t)=0$), and equals zero otherwise.

\subsection{Design of Constrained MDP}~\label{section_MDP_design}
As shown in Fig.~\ref{fig: time_relationship}, the processing latency can be larger than one time slot, the service rates allocated in the current time slot affect the available service rates and the DoS probability in the future time slots, which makes problem~\eqref{eq: formulated_optimization} a sequential decision-making problem. Since DRL is well-suited for solving Markovian problems, we resort to reformulating problem~\eqref{eq: formulated_optimization} as a constrained MDP. 

We first define the action, state, instantaneous reward and cost, and long-term reward and cost.
\subsubsection{Action}
The action to be taken in the $t$-th time slot is the service rate of the $i$-th BS, $a_{i}(t)$ (CPU cycles/slot), shown in problem~\eqref{eq: formulated_optimization}. We denote $\Delta f$ and $F$ as the minimum and maximum service rates that can be allocated. If the requests are denied, $a_i(t) = 0$. Otherwise, $a_i(t)$ could be any value between ${\Delta}f$ and $F$. Thus, the action space can be described as $\mathcal{A}= \{0\} \cup \mathcal{F}$ (CPU cycles/slot), where $\mathcal{F}=[\Delta f, F]$.

\subsubsection{State}
The required CPU cycles of the requests within {each} time slot are bounded by $f_{\text{r},\max} = \max\{f_\text{r}(t)\}$. Given the minimum service rate, $\Delta f$, the maximum processing latency can be expressed as
\begin{equation}
T_{\max}
=\frac{f_{\text{r},\max}}{\Delta f}.
\label{eq: T_max}
\end{equation}
We design the state of the $i$-th BS includes the states of all the service rates allocated in the past $T_{\max}$ time slots and can be described as 
\begin{equation} 
\begin{split} 
{\boldsymbol{s}}_{i}(t)
&= \{\hat{\boldsymbol{s}}_{i}(t,t'), t' \in [t-T_{\max},t]\} \\
&= \{\hat{\boldsymbol{s}}_{i}(t,t-T_{\max}),\cdots,\hat{\boldsymbol{s}}_{i}(t-1,t),\hat{\boldsymbol{s}}_{i}(t,t)\}, 
\label{eq: state_origin} 
\end{split} 
\end{equation}
where $\hat{\boldsymbol{s}}_{i}(t,t')=[\hat{\tau}_{i}(t,t'),\hat{a}_{i}(t,t')]$ is the state of the service rates allocated in the $t'$-th time slot, which is composed of the remaining processing latency and the service rates allocated in the $t'$-th time slot (See Appendix~\ref{appendix_singleState}). We denote $t$ as the current time slot, and $t'$ as the time slot that the service rate allocated, respectively. {We note} that both $t$ and $t'$ are integers, and $0 \leq t' \leq t$. {Therefore, the state space can be described as $\mathcal{S}=\{\boldsymbol{s}_i(t),  t \in [0,T_{\max}]\}$.}

\subsubsection{Instantaneous Reward and Cost}
The instantaneous reward is defined as
\begin{align} 
r_{i}(t)= 
\begin{cases} 
0,            & \text{if } a_{i}(t)=0 \\ 
-{r}_{i}^\text{a}(t),    & \text{if } a_{i}(t)\neq 0 \\ 
\end{cases} 
\label{eq: instantaneous_reward} 
\end{align} 
where {${r}_{i}^\text{a}(t)= {\tau_{i}(t)}/{\tau_{\max}}$} is the normalized processing latency when service rate is allocated, and $\tau_{\max}$ is the maximum value of $\tau_{i}(t)$ in~\eqref{eq_processing_latency}. We denote $\overline{{\ell}}_\text{b} = \ell_\text{h}+\ell_\text{c}\overline{\lambda}$ as the average size of the blocks, $\overline{\lambda}$ as the average number of arrived requests, and $\overline{\ell}_{u_\text{r}} $ as the average size of the requests in each time slot.

The instantaneous cost function is the BS DoS indicator function in the $t$-th time slot which is defined in~\eqref{eq_DoS_probability}.

\subsubsection{Long-Term Reward and Cost} 
Given a policy $\mu(\hat{s}_i(t))$, the long-term discounted reward, {representing the normalized processing latency when computation resources are allocated,} is defined as
\begin{equation} 
R_{i,\mu}(t)=\mathbb{E}_{\mu}\left [\sum\limits_{\hat{t}=t}^{\infty}  {\gamma_r^{\hat{t}-t} r_{i}(t)} \right], 
\label{eq: long_term_reward} 
\end{equation} 
where $\gamma_{r}$ is the reward discount factor. The long-term discounted cost, {representing the long-term DoS probability,} is defined as
\begin{equation} 
\begin{split}
C_{i,\mu}(t) 
= \mathbb{E}_{\mu}\left [\sum\limits_{\hat{t}=t}^{\infty}  {\gamma_c^{\hat{t}-t} c_{i}(t)} \right] 
=\dfrac{\mathbb{E}_{\mu}[c_{i}(t)]}{1-\gamma_{c}},
\label{eq: long_term_DoS} 
\end{split}
\end{equation} 
where $\gamma_c$ is the cost discount factor. To guarantee the requirement on the DoS probability, the long-term cost should satisfy the following constraint
\begin{equation} 
C_{i,\mu}(t) \leq 
\mathcal{E}_{\max},
\label{eq: DoS_constraint} 
\end{equation} 
where {$\mathcal{E}_{\max}={\epsilon_{\max}}/(1-\gamma_{c})$ is the maximum long-term DoS probability}, and $\epsilon_{\max}$ denotes the required threshold of the instantaneous DoS probability.

 \subsection{Reformulated Constrained MDP Problem}~\label{section_Lagran_alg}
Based on the above constrained MDP design parameters, we can reformulate problem~\eqref{eq: formulated_optimization} {(See Appendix~\ref{appendix_equivalence} for the proof of equivalence of problems~\eqref{eq: formulated_optimization} and~\eqref{eq: formulated_cmdp}.)} as a constrained MDP given by
\begin{align} 
\begin{split} 
&\max_{\mu(\cdot)}\quad  R_{i,\mu}(t)\\ 
&\begin{array}{r@{\quad}l@{}l@{\quad}l} 
\text{s.t.}\quad & C_{i,\mu}(t)\leq \mathcal{E}_{\max}.
\label{eq: formulated_cmdp} 
\end{array} 
\end{split} 
\end{align} 

To solve problem~\eqref{eq: formulated_cmdp}, we utilize a constrained DRL algorithm in which the policy, $\mu(\cdot)$, and the dual variable, $\lambda_{\mathcal{L}}$, are updated iteratively. The Lagrangian function of problem~\eqref{eq: formulated_cmdp} is given by
\begin{equation} 
\mathcal{L}_{i,t}\left(\mu(\cdot), \lambda_{\mathcal{L}}\right)=R_{i,\mu}(t)-\lambda_{\mathcal{L}}\left(C_{i,\mu}(t)-\mathcal{E}_{\max}\right), 
\label{eq: Lagrangian_function} 
\end{equation} 
where $\lambda_{\mathcal{L}}$ is the Lagrangian dual variable. Problem~\eqref{eq: formulated_cmdp} can be converted to the following unconstrained problem 
\begin{equation} 
\left(\mu^{*}(\cdot),\lambda_{\mathcal{L}}^*\right)=\arg \min\limits_{\lambda_{\mathcal{L}} \geq 0} \max\limits_{\mu(\cdot)} \mathcal{L}_{i,t}(\mu(\cdot),\lambda_{\mathcal{L}}), 
\label{eq: Lagrangian_optimization} 
\end{equation} 
where $\mu^{*}(\cdot)$ and $\lambda_\mathcal{L}^{*}$ indicate the optimal policy and the optimal Lagrangian dual variable, respectively. To apply the constrained DRL algorithm, we further verify that the formulated problem in eq.~\eqref{eq: formulated_cmdp} satisfies the Markov property~\cite{BOOK_SCY_drl} (See proof in Appendix~\ref{appendix_MarkovProperty}).

\subsection{Service Rate Allocation} 

To solve the coupled MDP problem formulated in eq.~\eqref{eq: formulated_cmdp}, we decouple this problem by utilizing the PD-DDPG algorithm to find the optimal primal-dual solution~\cite{Lagrangian_PrimalDual}. We define the reward critic Q-network as $Q_{R}(\widetilde{\boldsymbol{s}},a|\theta_R)$, the cost critic Q-network as $Q_{C}(\widetilde{\boldsymbol{s}},a|\theta_C)$, the actor network as $\mu(\widetilde{\boldsymbol{s}}|\theta_{\mu})$, respectively. The corresponding target critic networks are $Q'_{R}(\widetilde{\boldsymbol{s}},a|\theta'_R)$, $Q'_{C}(\widetilde{\boldsymbol{s}},a|\theta'_C)$, and $\mu'(\widetilde{\boldsymbol{s}}|\theta'_{\mu})$. The experience replay memory buffer as $\mathcal{B}$. The specific algorithm is detailed in Algorithm~\ref{Algorithm: primal_dual_ddpg}\footnote{Since the steps always refer to the $i$-th BS, we do not explicitly denote ``$i$'' in this algorithm.}.

\begin{algorithm}[t] 
\algsetup{linenosize=\normalsize} \normalsize  
\caption{Dynamic {Resource} Allocation {Algorithm}} 
\label{Algorithm: primal_dual_ddpg} 
Randomly initialize $Q_{R}(\widetilde{\boldsymbol{s}},a|\theta_R)$, $Q_{C}(\widetilde{\boldsymbol{s}},a|\theta_C)$, and $\mu(\widetilde{\boldsymbol{s}}|\theta_{\mu})$.
Initialize $\theta'_R \gets  \theta_R$, $\theta'_C \gets  \theta_C$, and $\theta'_{\mu}\gets \theta_{\mu}$, $\lambda_{\mathcal{L}}$, $\epsilon_{\max}$, $\mathcal{B}$, $\gamma_r$, and $\gamma_c$.\\ 
{ 
{ 
Select action based on the current policy $\theta_{\mu}$ and the exploration noise $\mathcal{N}_{a}(t)$ as: 
$\quad\quad\quad\quad\quad\quad a(t) =\mu(\widetilde{\boldsymbol{s}}(t)|\theta_{\mu}) + \mathcal{N}_{a}(t). \nonumber $

Execute action $a_{i}(t)$ then observe reward $r(t)$, cost $c(t)$, and new state $\widetilde{\boldsymbol{s}}(t+1)$.\\ 
Store transition $\langle \widetilde{\boldsymbol{s}}(t),a(t),r(t),c(t),\widetilde{\boldsymbol{s}}(t+1) \rangle$ into $\mathcal{B}$. 
\\ 
Randomly sample a mini-batch of $M$ transition from $\mathcal{B}$: \{$\langle \widetilde{\boldsymbol{s}}_m,a_m,r_m,c_m,\widetilde{\boldsymbol{s}}_{m+1} \rangle$, $m=1,2,\cdots,M$ \}.\\ 
Set Q-targets for reward and cost temporal difference (TD): 
$y_m^{R} = r_m + \gamma_r Q'_R\left(\widetilde{\boldsymbol{s}}_{m+1}, \mu '(\widetilde{\boldsymbol{s}}_{m+1}|\theta'_{\mu})|\theta'_R\right), \quad \quad$
$y_m^{C} = c_m + \gamma_c Q'_C\left(\widetilde{\boldsymbol{s}}_{m+1}, \mu '(\widetilde{\boldsymbol{s}}_{m+1}|\theta'_{\mu})|\theta'_C\right).$

Update reward and cost critic Q-networks by minimizing the losses denoted by mean squared TD errors: 
$\Delta_R = \dfrac{1}{M} \sum\limits_{m}^{}\left(y_{m}^{R} - Q_R (\widetilde{\boldsymbol{s}}_{m},a_{m}|\theta_R)\right)^2,\quad\quad\quad$
$\Delta_C = \dfrac{1}{M} \sum\limits_{m}^{}\left(y_{m}^{C} - Q_C (\widetilde{\boldsymbol{s}}_{m},a_{m}|\theta_C)\right)^2. $

Update the actor policy $\theta_{\mu}$, in the primal domain, with sampled policy gradient descent: 
$\quad\quad\quad\quad\quad\quad\quad\quad\quad\quad\quad
\nabla_{\theta_{\mu}} \mathcal{L} 
=  \dfrac{1}{M} \sum\limits_{m}^{}   
\nabla_{\theta_{\mu}} 
(Q_R\left(\widetilde{\boldsymbol{s}}_m, \mu(\widetilde{\boldsymbol{s}}_m|\theta_{\mu})|\theta_R\right) 
 - \lambda_{\mathcal{L}} Q_C(\widetilde{\boldsymbol{s}}_m,\mu(\widetilde{\boldsymbol{s}}_m|\theta_{\mu})|\theta_C)).  $

Calculate the gradient of dual variable $\lambda_{\mathcal{L}}$:
$\nabla_{\lambda_{\mathcal{L}}} \mathcal{L} 
= \dfrac{1}{M} \sum\limits_{m}^{}\left(Q_C(\widetilde{\boldsymbol{s}}_m,\mu(\widetilde{\boldsymbol{s}}_m|\theta_{\mu})|\theta_{C})-{\mathcal{E}_{\max}}
\right).$

Update the dual variable, $\lambda_\mathcal{L}$, in the dual domain, with sampled dual gradient ascent: 
$\lambda_{\mathcal{L}} \gets \max\{0, \lambda_{\mathcal{L}}+\alpha_{\lambda_\mathcal{L}}\nabla_{\lambda_{\mathcal{L}}} \mathcal{L}(\theta_{\mu},\lambda_{\mathcal{L}})\}. $

Update target networks with $\varphi$: 
$\theta'_{R}       \gets \varphi \theta_{R}      + (1-\varphi)\theta'_{R}, \quad\quad\quad\quad\quad\quad\quad$
$\theta'_{C}       \gets \varphi \theta_{C}      + (1-\varphi)\theta'_{C}, \quad\quad\quad\quad\quad\quad\quad$
$\theta'_{\mu} \gets \varphi \theta_{\mu} + (1-\varphi)\theta'_{\mu}. $
} 
} 
\end{algorithm} 

We consider that the agent follows a deterministic policy denoted by $\mu: a(t)=\mu(\boldsymbol{s}(t)|\theta_{\mu})$. It is a neural network that determines the service rates allocation based on the state in each time slot, where $\theta_{\mu}$ represents the parameter of the neural network.

To improve the training efficiency of the constrained DRL algorithm, we propose to reduce the dimension of the state space of the constrained MDP defined in eq.~\eqref{eq: state_origin}  by extracting key features to achieve a lower dimension state for training. Specifically, we extract remaining processing latency, $\hat{\tau}_{i}(t,t')$, of all the allocated service rates in the $i$-th BS to be a normalized sum, i.e.,
\begin{equation} 
\begin{split} 
\rho_{i}(t) 
=\dfrac{\tau_{i,\min}(t)}{\tau_{\max}}
= \dfrac{{\dfrac{1}{F}\left(\sum\limits_{t'=t-{T_{\max}}}^{t} \hat{\tau}_{i}(t,t')\right)} } 
{\tau_{\max}},
\label{eq: workload} 
\end{split} 
\end{equation} 
where $\tau_{i,\min}(t)$ is the remaining processing latency of all the allocated service rates if they are processed with the maximum service rate, $F$. Based on eq.~\eqref{eq: workload}, the lower dimension state is re-designed as
\begin{equation} 
\begin{split} 
\widetilde{\boldsymbol{s}}_{i}(t)=\left\{\dfrac{f_{i,\text{a}}(t)}{F},\rho_{i}(t)\right\}, 
\label{eq: feature_normalization} 
\end{split} 
\end{equation} 
where $f_{i,\text{a}}(t)$ is the available service rates of the $i$-th BS in the $t$-th time slot.

We further apply transfer learning to reduce the training time due to changes in the DoS probability constraints. To do so, we reuse the parameters of well-trained neural networks as the initial parameters for the target neural network.

\subsection{Complexity Analysis}
{The computational complexity of the proposed constrained DDPG composes the inference complexity of three neural networks denoted by $Q_{R}(\widetilde{\boldsymbol{s}}, a|\theta_R)$, $Q_{C}(\widetilde{\boldsymbol{s}}, a|\theta_C)$, and $\mu(\widetilde{\boldsymbol{s}}|\theta_{\mu})$, respectively. Thus, the computational complexity of the proposed constrained DDPG algorithm is 
\begin{equation}
    O_\mathrm{PRO} = O(N_\mathrm{R} + N_\mathrm{C} + N_{\mu}),
    \label{eq_complexity_proposed}
\end{equation}
where $N_\mathrm{R}$, $N_\mathrm{C}$, and $N_{\mu}$ are the number of multiplications required to process the three neural networks, and are given by $N_\mathrm{R} = \sum_{\ell_\mathrm{R}=1}^{L_\mathrm{R}} n_{\ell_\mathrm{R}} n_{\ell_\mathrm{R}+1}$, 
$N_\mathrm{C} = \sum_{\ell_\mathrm{C}=1}^{L_\mathrm{C}} n_{\ell_\mathrm{C}} n_{\ell_\mathrm{C}+1}$, 
and $N_{\mu} = \sum_{\ell_{\mu}=1}^{L_{\mu}} n_{\ell_{\mu}} n_{\ell_{\mu}+1} + \Omega_\mathrm{Sig}$, 
respectively, where $L_\mathrm{R}$, $L_\mathrm{C}$, and $L_{\mu}$ denote the number of layers in the neural networks; $n_{\ell_\mathrm{R}}$, $n_{\ell_\mathrm{C}}$, and $n_{\ell_{\mu}}$ are the number of neurons in the $\ell_\mathrm{R}$-th, $\ell_\mathrm{C}$-th, and $\ell_{\mu}$-th layer, and $\Omega_\mathrm{Sig}$ represents the number of multiplications required to compute the $Sigmoid$ function. We note that the computation complexity of $ReLU$ function is ignored since it's very low compared with the other operations.}

{Another commonly used method for solving sequential decision-making problems is dynamic programming, whose computational complexity is given by
\begin{equation}
    O_\mathrm{DP} = O(|\mathcal{A}| \times |\mathcal{S}| \times N_\mathrm{iter} ),
    \label{eq_complexity_DP}
\end{equation}
where $|\mathcal{A}|$ and $|\mathcal{S}|$ represent the size of the action and state spaces given in Section~\ref{section_MDP_design}, and $N_\mathrm{iter}$ is the number of iterations required for the dynamic programming algorithm to converge. Since our problem involves continuous action space, where the action can take any value ranging from $f$ to $F$ or equal $0$, $|\mathcal{A}|$ approaches infinity, resulting in $O_\mathrm{DP}$ also approaching infinity. Consequently, conventional dynamic programming is not a feasible option for solving our problem shown in eq.~\eqref{eq_complexity_DP}.}

\begin{table}[t] 
\renewcommand\arraystretch{1} 
\caption{Key Simulation Parameters} 
\centering 
\begin{tabular}{m{5cm} l}
\toprule 
\toprule 
\textbf{Simulation parameters}&\textbf{Values}\\
\toprule 
Number of overall BSs in the network $N_\text{B}$                                                    &10\\ 
\hline 
Computation capacity of the BSs $F$                                                                  &1.6 G CPU cycles/slot\cite{SCY_digital_twin}\\
\hline
Minimum service rate can be allocated $\Delta f$                                                     &0.01 G CPU cycles/slot\\
\hline
Request arrival rate $\bar{\lambda}$                                                                 &$\textrm{Poisson}$($10^3$)\\ 
\hline 
Size of request by the $u$-th user ${\ell}_{u}(t)$                                                   &Uniform$(1, 10)$ KB\\
\hline
Coefficient size of user requests $\ell_\text{c}$                                                    &8 bytes\\ 
\hline 
Prior probability of event $e_i(t)$                                                                  &0.8\\ 
\hline 
{Weight coefficient of reputation threshold $\eta$}                                             &1\\ 
\hline 
Coefficient of reputation $\vartheta_\mathrm{I}$                                                     &0.2\\ 
\hline 
Coefficient of CPU cycles for blockchain ${\kappa}_\text{bc}$                                        &0.001 G~\cite{RichardYu_MDP_SNRstate}\\
\hline 
Coefficient of CPU cycles for services ${\kappa}_\text{sp}$                                          &330~\cite{SCY_digital_twin}\\ 
\hline 
Threshold of DoS probability $\epsilon_{\max}$                                                       &2 \%          \\ 
\bottomrule 
\bottomrule 
\end{tabular} 
\label{tab: Simulation Parameters} 
\end{table} 

\begin{figure}[t]
\centering 
\subfigure[Reputations evaluated with different $\overline{\lambda}$ when $\Pr\{e_i(t)\}=0.8$.]
{\includegraphics[height=6.5cm, width=9cm]{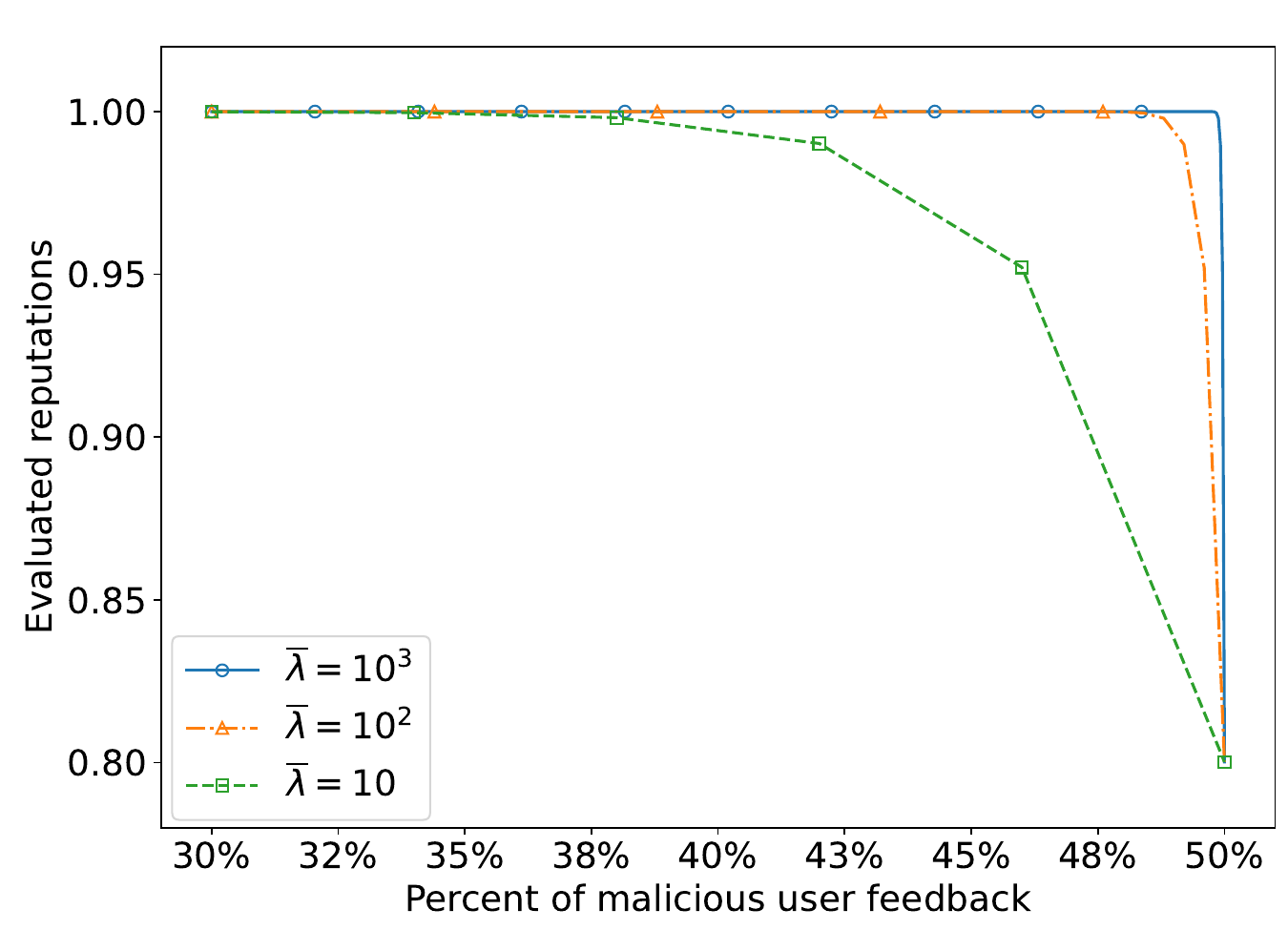}\label{fig_reputation_Bayesian_packet}}
\subfigure[Reputations evaluated with different $\Pr\{e_i(t)\}$, when $\overline{\lambda}=100$.]
{\includegraphics[height=6.5cm, width=9cm]{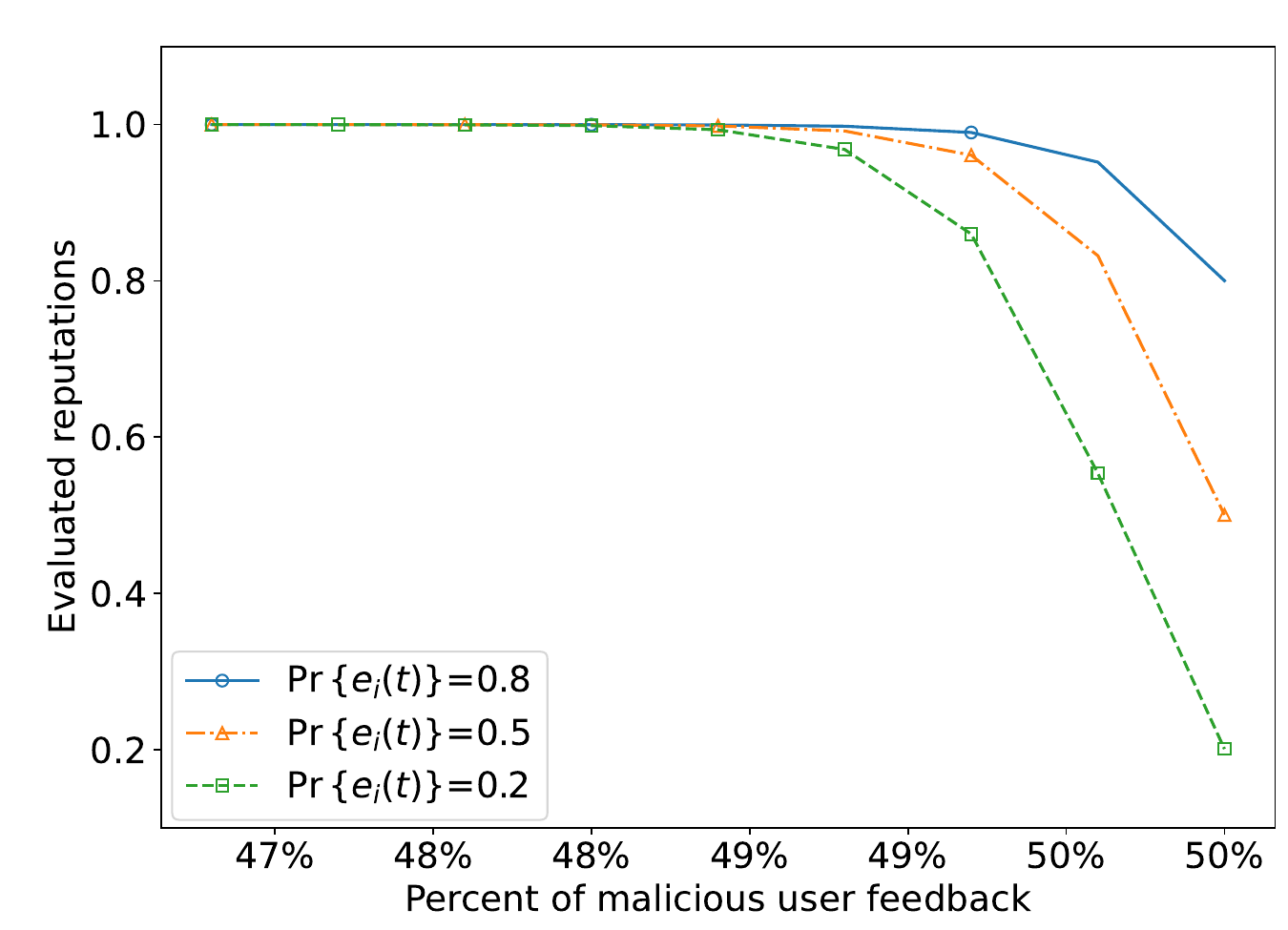}\label{fig_reputation_Bayesian_prior}}
\caption{Evaluated BS reputations {under malicious user feedback attacks}.}
\label{fig_reputations}
\end{figure}
\section{Simulations and Empirical Convergence Analysis}
\subsection{Simulation Setup}~\label{sec_sim_setup}
We present numerical simulations to evaluate the performance {and analyze the empirical convergence} of our proposed BC-DRL solution using Google TensorFlow embedded in a Python platform. We consider a total of 10 BSs that are initialized as non-malicious and have the maximum reputations equal to one. The transmission rates among all the BSs are assumed to be $W$=10Gbps~\cite{BS_10Gbps}. The action exploration noise in our DRL simulations follows an Ornstein-Uhlenbeck process~\cite{ddpg_Google_deepMind}. We note that the service rates of the committee BSs are all considered equal to $a_i(t)$ in the $t$-th time slot, since the serving BS is randomly assigned from the committee in each time slot. Unless otherwise mentioned, the simulation parameters are summarized in Table~\ref{tab: Simulation Parameters}.

\subsection{Simulation Results}
\subsubsection{RPoS-Based Blockchain Management}
We present simulations of the reputation evaluation and performance {of} our RPoS consensus protocol.

Fig.~\ref{fig_reputations} {explores the impact brought by malicious user feedback attacks discussed in Section~\ref{section_malicious_user_feedback_attack} by evaluating the reputations evaluated based on Bayesian inference when the percentage of feedback from malicious users increases}. We see in Figs.~\ref{fig_reputations} (a) and (b) that the evaluated reputations decrease with increasing malicious user feedback. Fig.~\ref{fig_reputation_Bayesian_packet} shows that our Bayesian-based reputation evaluation with a larger user request rate, $\bar{\lambda}$, can accurately maintain the BS reputations to be equal to 1 even with a high percentage of malicious user feedback that is close to 50\%. In Fig.~\ref{fig_reputation_Bayesian_prior}, we see that a higher prior probability of requests being served by the BSs, $\Pr\{e_i(t)\}$, corresponds to a higher resistance ability to malicious user feedback attacks due to a higher confidence in the reputation evaluation of the BSs.

\begin{figure}[t]
\centering 
\subfigure[CPU cycles required {by the blockchain miner BS}. A lower CPU cycle requirement corresponds to a higher resource efficiency and lower BS DoS probability.]
{\includegraphics[height=6.5cm, width=9cm]{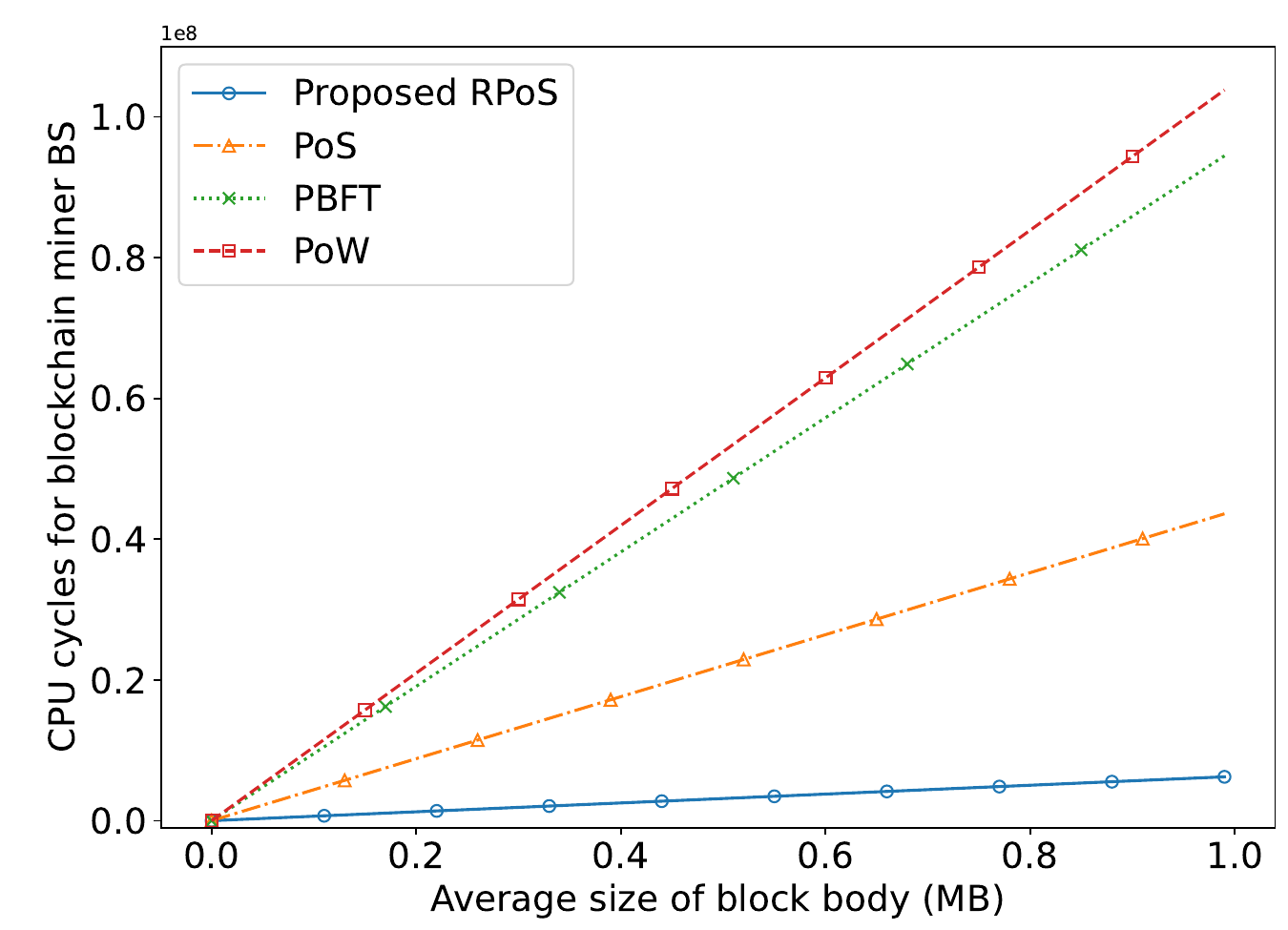}\label{fig_computation_require}}
\subfigure[{Tampering time for the optimized parameters of the neural networks with different system management scenarios.}]
{\includegraphics[height=6.5cm, width=9cm]{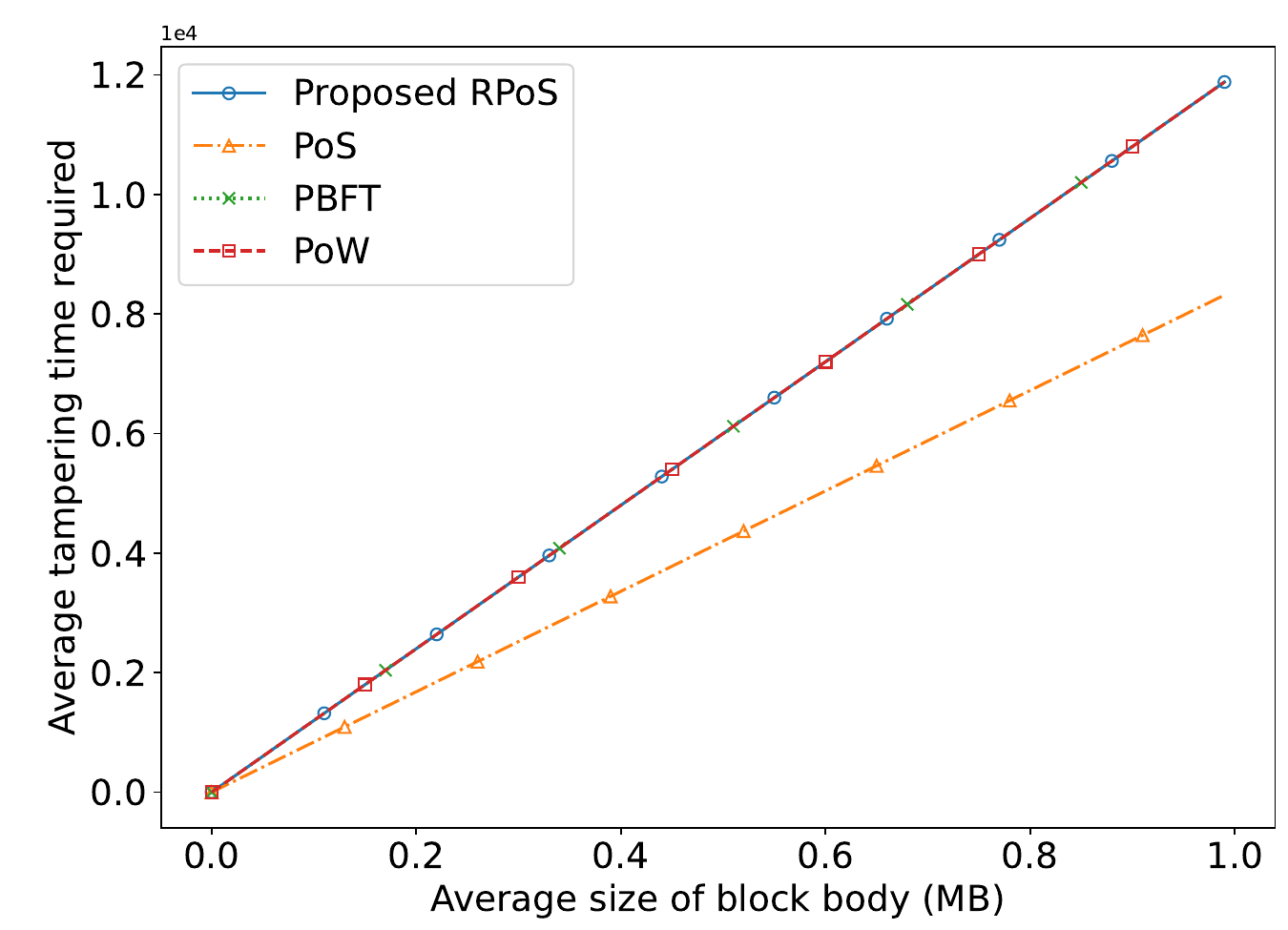}\label{fig_tamper_time}}
\caption{{Computation resource consumption and tampering resistant ability for different blockchain consensus protocols.}}
\label{fig_attacks}
\end{figure}

{Fig.~\ref{fig_attacks} shows the computation resource consumption and tampering resistant ability of different consensus when there are three malicious BSs.} {We can observe from Fig.~\ref{fig_computation_require}} that {the proposed} RPoS requires significantly lower CPU cycles compared to other benchmark consensus protocols. {Fig.~\ref{fig_tamper_time} shows the required time to tamper the blockchain. This observation, combined with the findings presented in Fig.~\ref{fig_computation_require}, confirms that PoS is vulnerable to tampering attacks despite its resource efficiency. In contrast, the proposed RPoS consensus exhibits robust resistance to tampering attacks, comparable to the security levels achieved by PoW, while significantly reducing computation resource requirements. We can also observe from Fig.~\ref{fig_attacks} that both the CPU cycles for blockchain miner BS and tampering time increase linearly with the increasing average size of the block body. This linear relationship is attributed to the proportionality of these metrics to the block size, primarily determined by the block body size.}

\begin{table}[t] 
\renewcommand\arraystretch{1} 
\caption{Hyper-Parameters of DRL Algorithm} 
\centering 
\begin{tabular}{l l}
\toprule 
\toprule 
\textbf{Simulation parameters}&\textbf{Values}\\ 
\toprule 
Discount factors of rewards $\gamma_\text{r}$ , $\gamma_\text{c}$                                                                     &0.95\\ 
\hline
Learning rates of critic NNs $\alpha_\text{r}$, $\alpha_\text{c}$                                              &$5\times 10^{-4}$\\ 
\hline 
Learning rate of actor NN $\alpha_\text{a}$                                                                    &$2\times 10^{-4}$\\ 
\hline 
Learning rate of dual variable $\alpha_{\lambda_\mathcal{L}}$                                                  &$0.1$\\ 
\hline 
Mini-batch size $M$                                                                                            &512\\ 
\hline 
Target NNs updating rate $\varphi$                                                                             &$5\times 10^{-3}$\\ 
\hline
Max replay buffer size                                                                                         &$2\times 10^5$\\
\hline
{Number of steps in each episode $N_\mathrm{step}$}                                                       &{$1000$}\\
\bottomrule 
\bottomrule 
\end{tabular} 
\label{tab_HyperParameters} 
\end{table}

\subsubsection{DRL-{B}ased Resource Allocation}~\label{section_sim_DRL}
We compare the performance of our proposed algorithm with other benchmark dynamic resource allocation algorithms from the perspective of minimizing the processing latency and DoS probability, and improving the training efficiency. A user feedback value of ``1'' either indicates a BS refusing to provide services or malicious users sending incorrect feedback when service rates are allocated. Unless otherwise mentioned, the hyper-parameters are summarized in Table~\ref{tab_HyperParameters}.

\begin{figure}[t]
\centering 
\subfigure[Normalized processing latency when resources are allocated.]  
{\includegraphics[height=6.5cm, width=9cm]{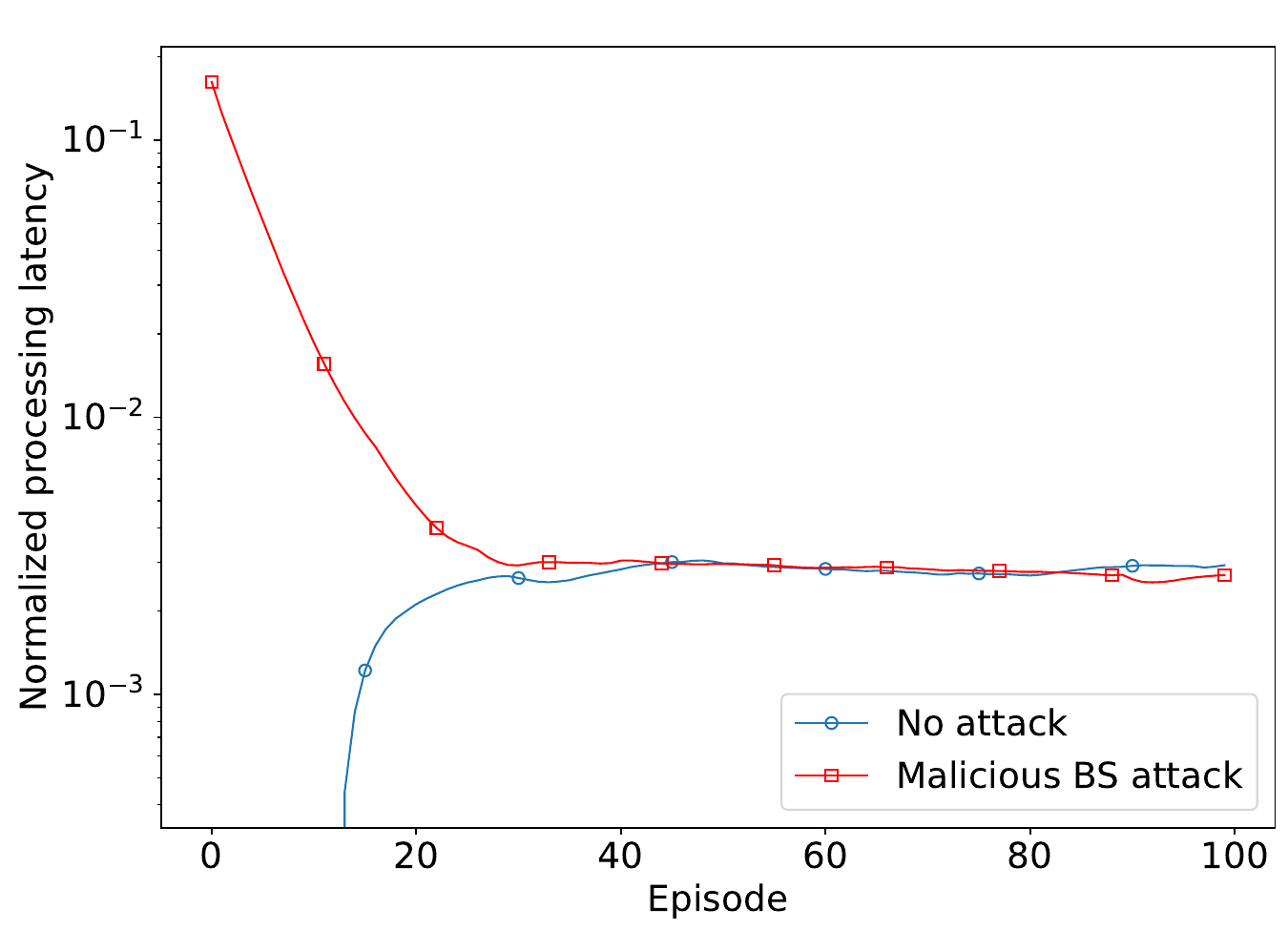}\label{fig: Training_results_reward_attack}} 
\subfigure[Average DoS probability.]
{\includegraphics[height=6.5cm, width=9cm]{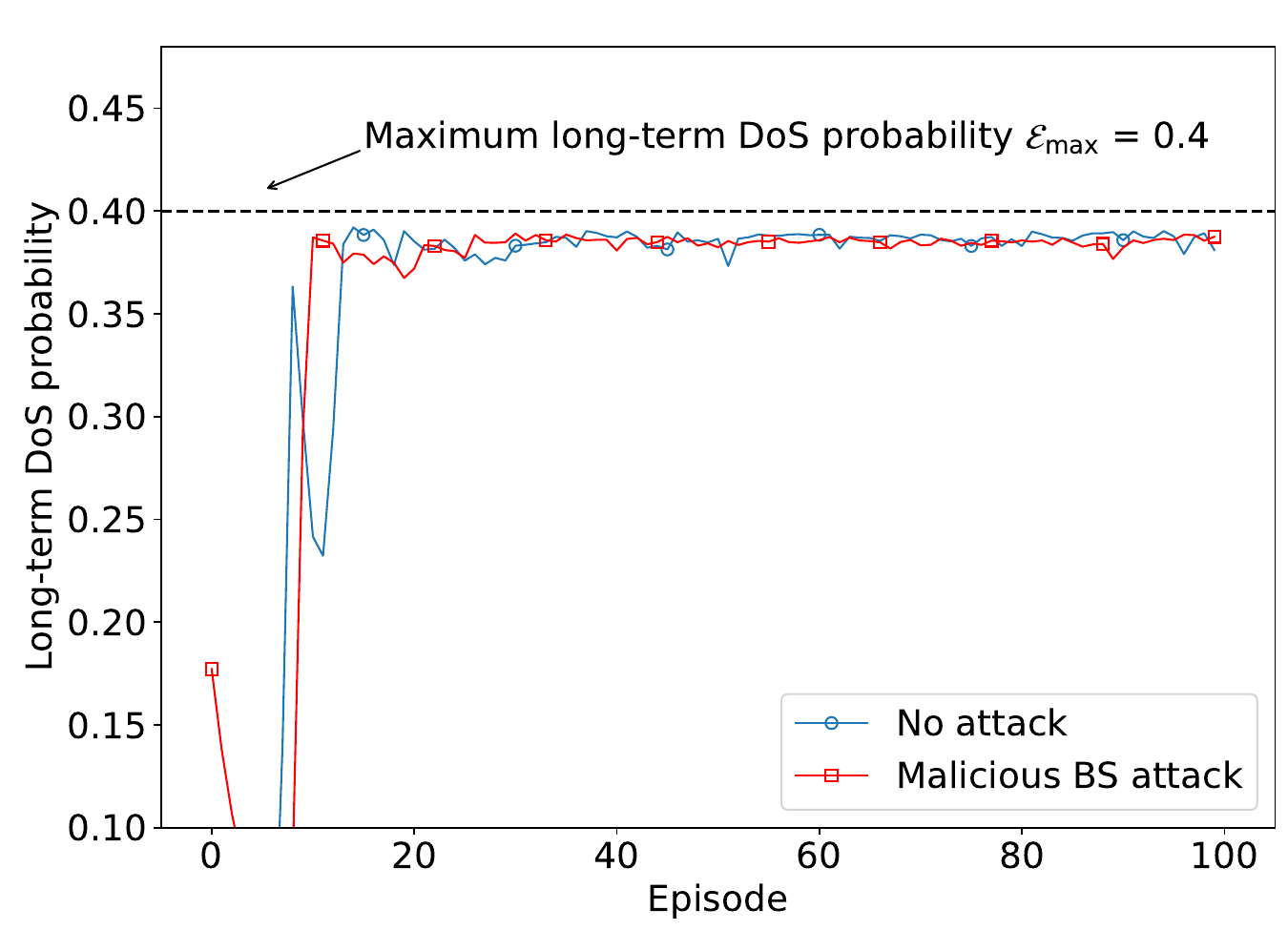}\label{fig: Training_results_cost_attack}} 
\caption{Performances of BC-DRL solution with and without {malicious BS attacks}.} 
\label{fig_training_results_attack}
\end{figure}
Fig.~\ref{fig_training_results_attack} shows the training results of BC-DRL with and without malicious BS attacks. {To investigate the impact of malicious BS attacks on processing latency and DoS probability, we include three malicious BSs in our simulation. Each BS randomly denies service requests from users following the Bernoulli process. We can observe that both the processing latency and the DoS probability converged to steady values after approximately 15 episodes.} We observe that while BC-DRL achieves approximately the same reward and cost performance for both scenarios of with and without malicious BSs, the dual variables for the optimization converge to different values in each scenario to accurately balance the trade-off between processing latency and DoS probability.

\begin{figure}[t]
\centering 
\subfigure[{Normalized processing latency when resources are allocated.}]  
{\includegraphics[height=6.5cm, width=9cm]{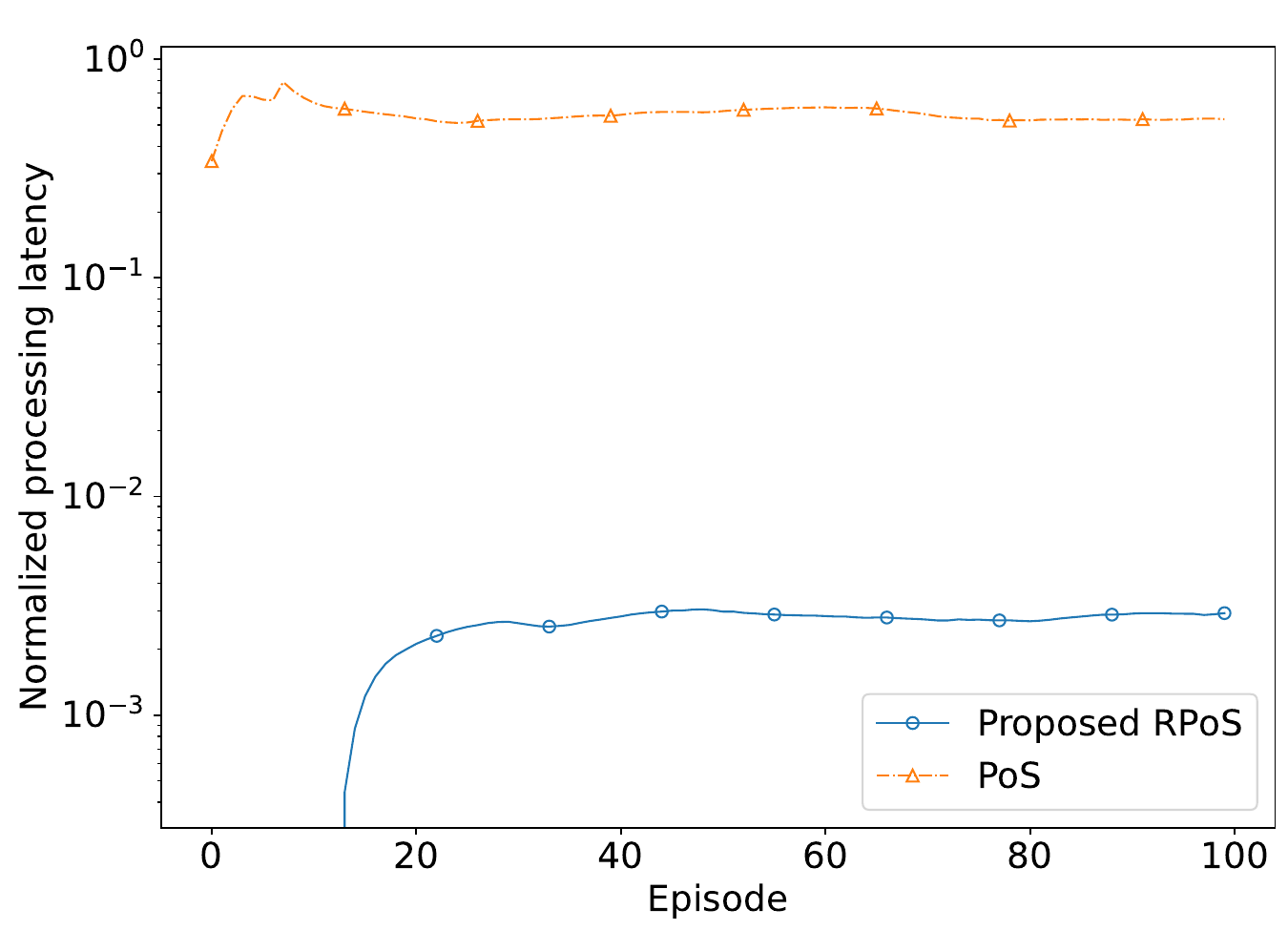}\label{fig: Training_results_reward_consensus}} 
\subfigure[{Average DoS probability in the long term.}] 
{\includegraphics[height=6.5cm, width=9cm]{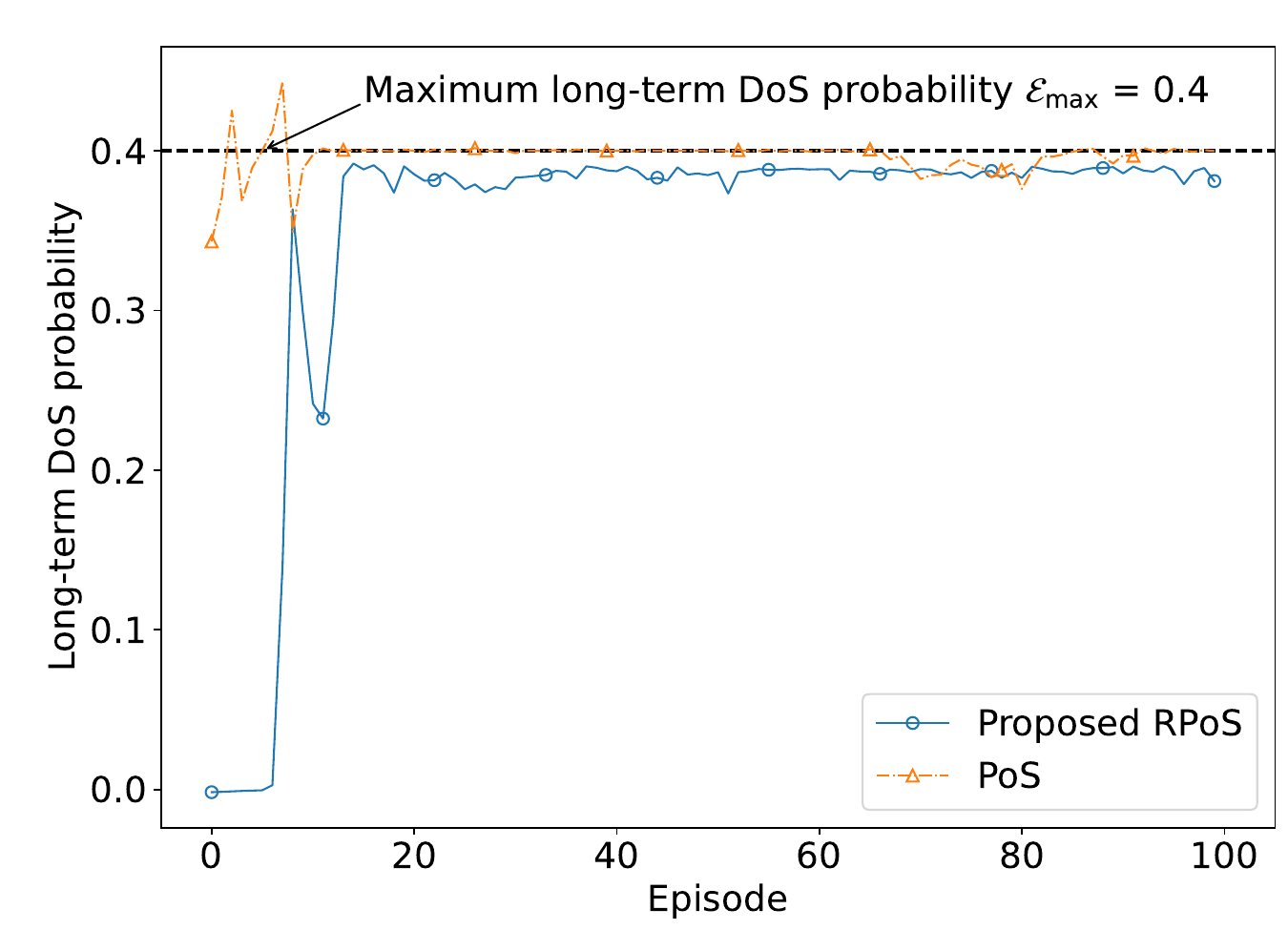}\label{fig: Training_results_cost_consensus}} 
\caption{{Performance of processing latency and DoS probability, when resource allocated by the proposed constrained DDPG algorithm}.} 
\label{fig: training_results_consensus}
\end{figure}

\begin{figure}[t]
\centering 
\subfigure[Normalized processing latency when resources are allocated.]
{\includegraphics[height=6.5cm, width=9cm]{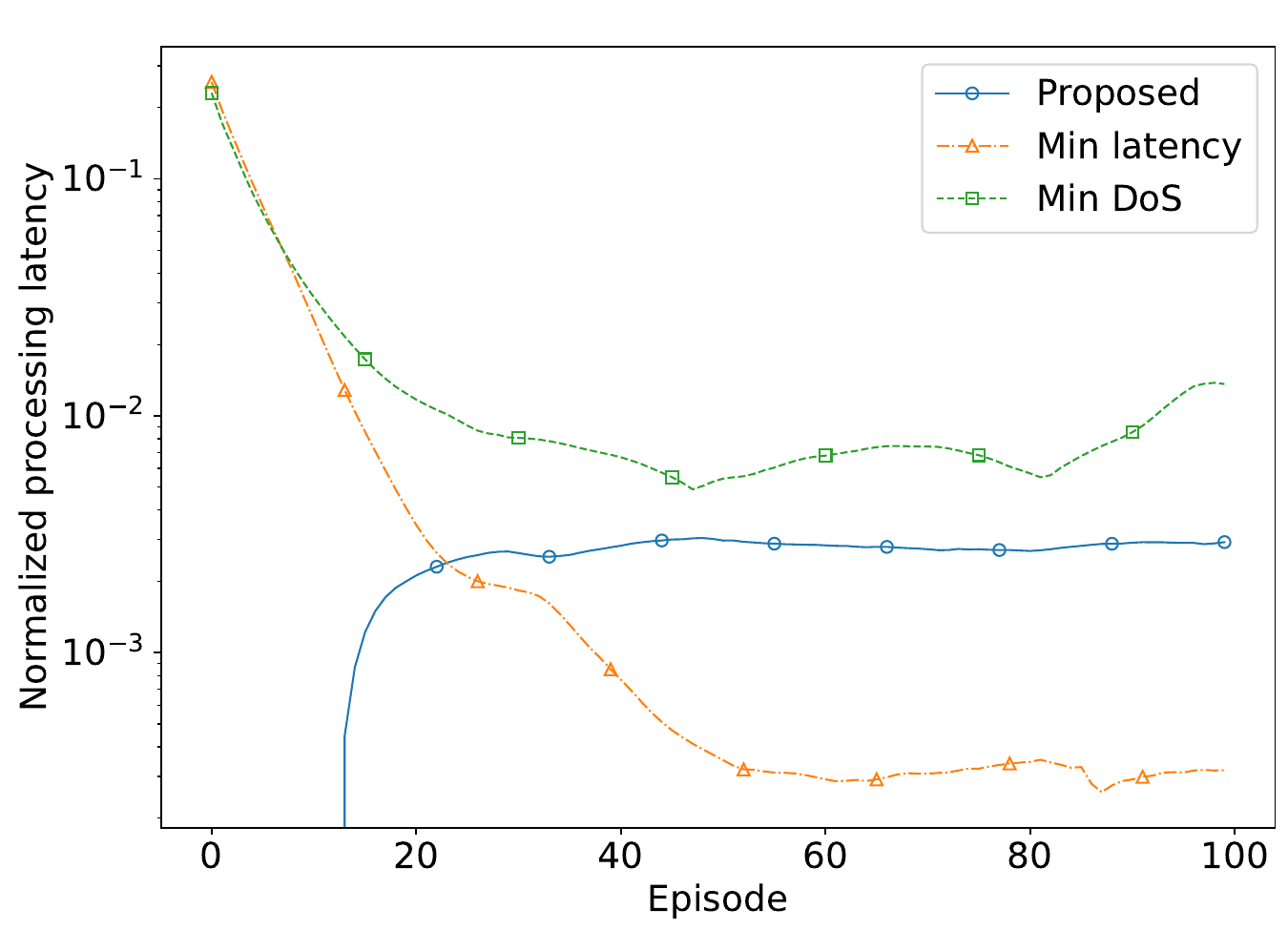}\label{fig: train_resource_allocation_reward}} 
\subfigure[Average DoS probability in the long-term.]
{\includegraphics[height=6.5cm, width=9cm]{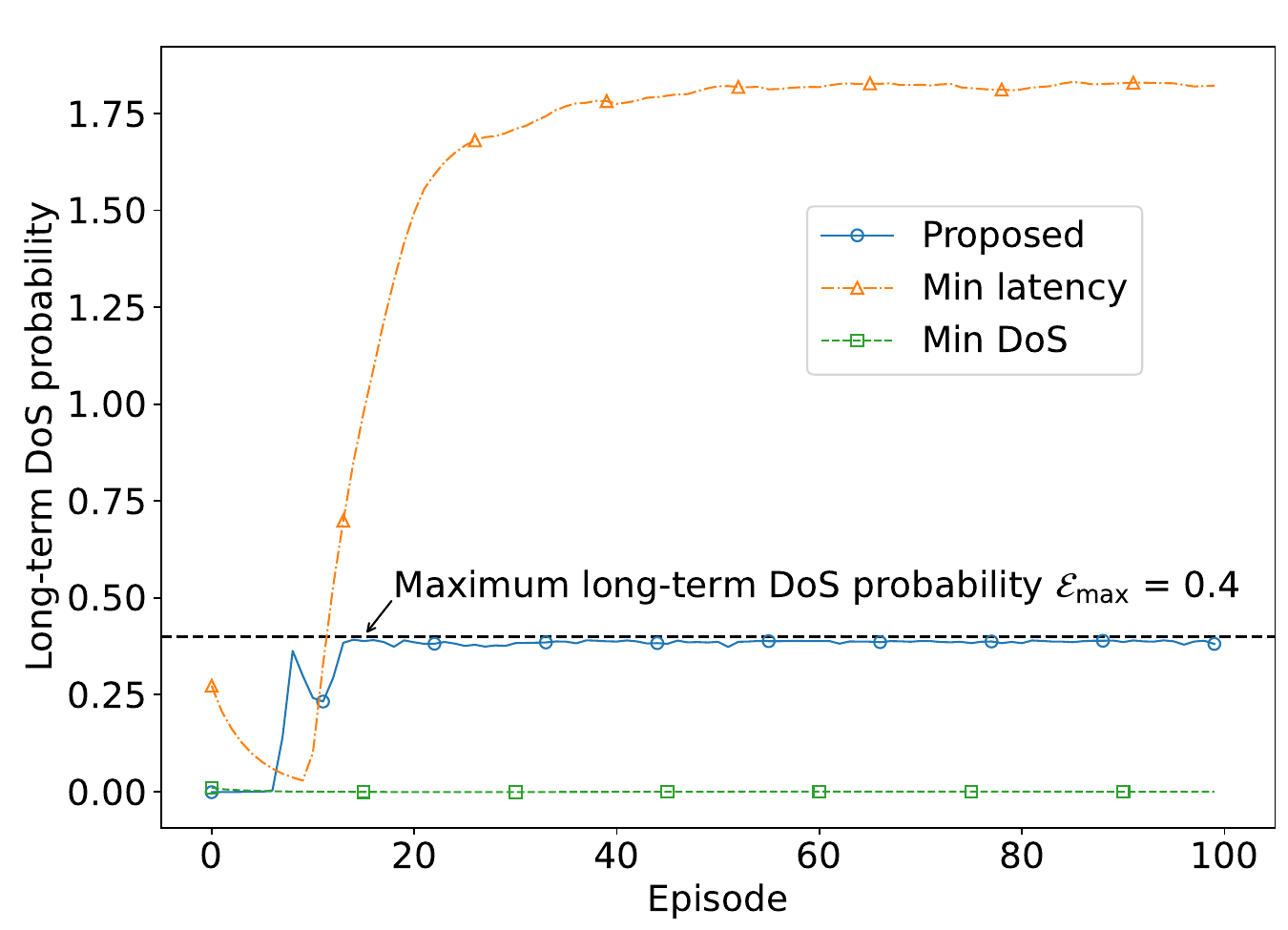}\label{fig: train_resource_allocation_cost}} 
\caption{{P}rocessing latency and DoS probability when resources are allocated with different DRL algorithms. Our proposed algorithm achieves minimum processing latency {with a satisfactory DoS probability.}}
\label{fig: training_results_resource_allocation}
\end{figure}
Fig.~\ref{fig: training_results_consensus} shows the training results of our proposed BC-DRL solution and a benchmark PD-DDPG resource allocation algorithm with {PoS} consensus. {The comparison between the RPoS and PoS consensus protocols shows that both protocols can satisfy the constraint requirements. However, the RPoS outperforms PoS in terms of achieving a lower processing latency. The superiority of RPoS can be attributed to its ability to allocate more computation resources to provide MEC services. This advantage is due to the random selection of the BS from the committee in the RPoS consensus protocol, allowing for a more balanced distribution of resources among the participating BSs. In contrast, the PoS protocol consistently chooses the same BS, which may result in uneven resource allocation and higher processing latency.}

In Fig.~\ref{fig: training_results_resource_allocation}, we compare our constrained DDPG DRL solution with benchmark unconstrained DDPG DRL solutions aimed at minimizing either the processing latency or DoS probability. {Figs.~\ref{fig: train_resource_allocation_reward} and~\ref{fig: train_resource_allocation_cost}} highlight a fundamental performance trade-off where the min latency solution leads to an intolerable DoS probability, whilst the min DoS probability solution results in a lengthy processing latency and an unnecessarily low DoS probability. {In contrast, our proposed constrained DRL solution can achieve a significantly reduced processing latency while maintaining a satisfactory DoS probability, as determined by the specified maximum long-term cost.}

\subsection{Empirical Convergence Analysis}
Empirical convergence analysis of our proposed constrained DRL is provided since the convergence of the proposed DRL is highly affected by the iterative updates and interactions with the MEC network. Specifically, we focus on analyzing the convergence ability with dynamic values of the DoS probability constraint, $\mathcal{E}_{\max}$.

\begin{figure}[t]
    \centering
    \includegraphics[height=6.5cm, width=9cm]{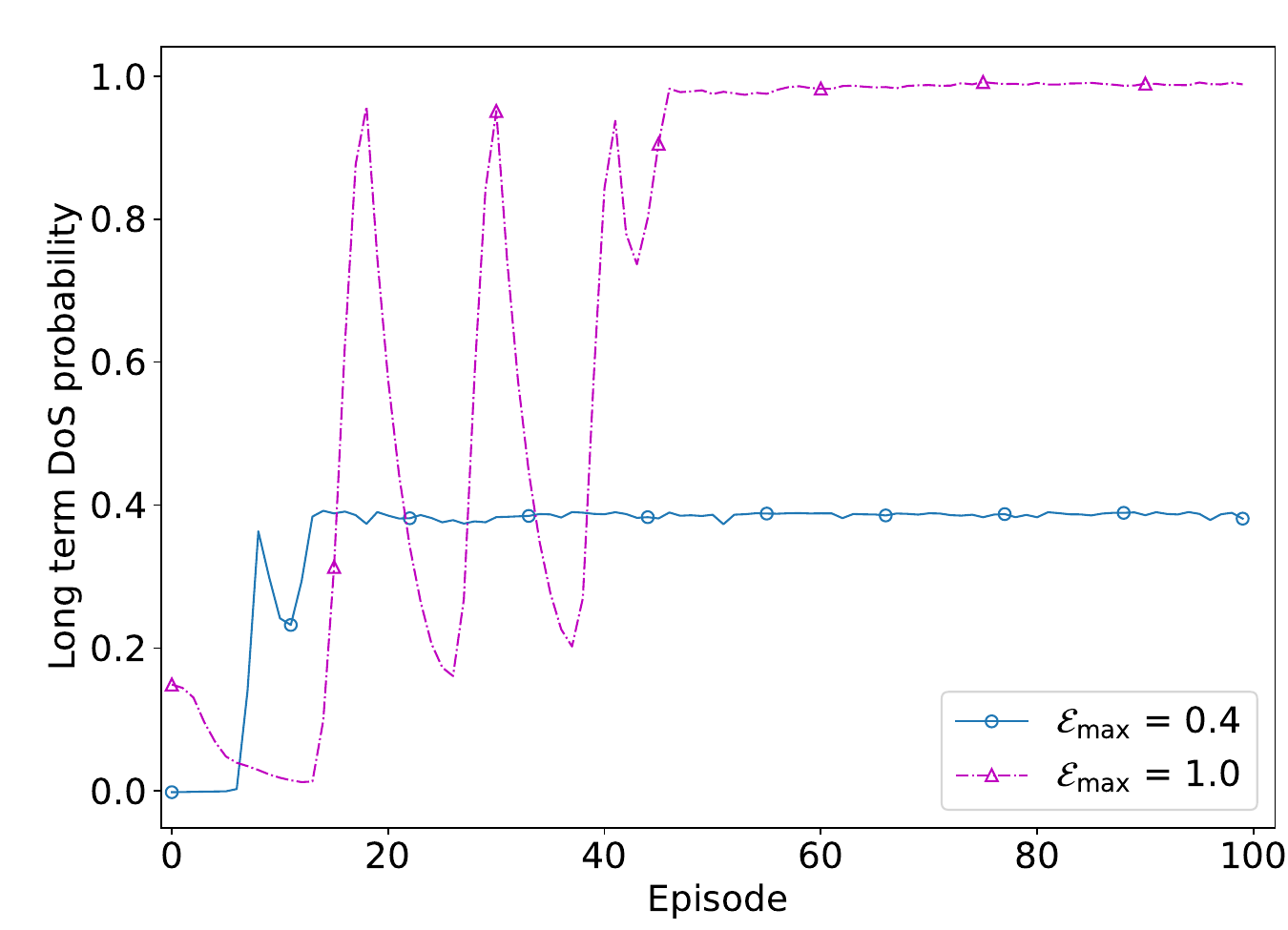}    
    \caption{{Training results of long-term DoS probability achieved by random initialization.}}  
    \label{fig_transfer_cost_random_init}
\end{figure}
Fig.~\ref{fig_transfer_cost_random_init} shows the training results of long-term DoS probability achieved by random initialization when the constraints on the DoS probability equal $0.4$ and $1.0$, respectively. This figure shows that both DoS probabilities converge to stable values with the increasing number of training episodes, indicating that the proposed algorithm is learning and optimizing the resource allocation efficiently. We can also observe from this figure that it takes approximately $15$ episodes to converge when the constraint on the DoS probability is $0.4$, whilst the number of episodes before convergence increases to approximately $47$ when $\mathcal{E}_{\max}$ increases to $1.0$. This phenomenon indicates that the difference in the constraint on DoS probability significantly affects the convergence speed of the proposed algorithm.

\begin{figure}[t]
    \centering
    \includegraphics[height=6.5cm, width=9cm]{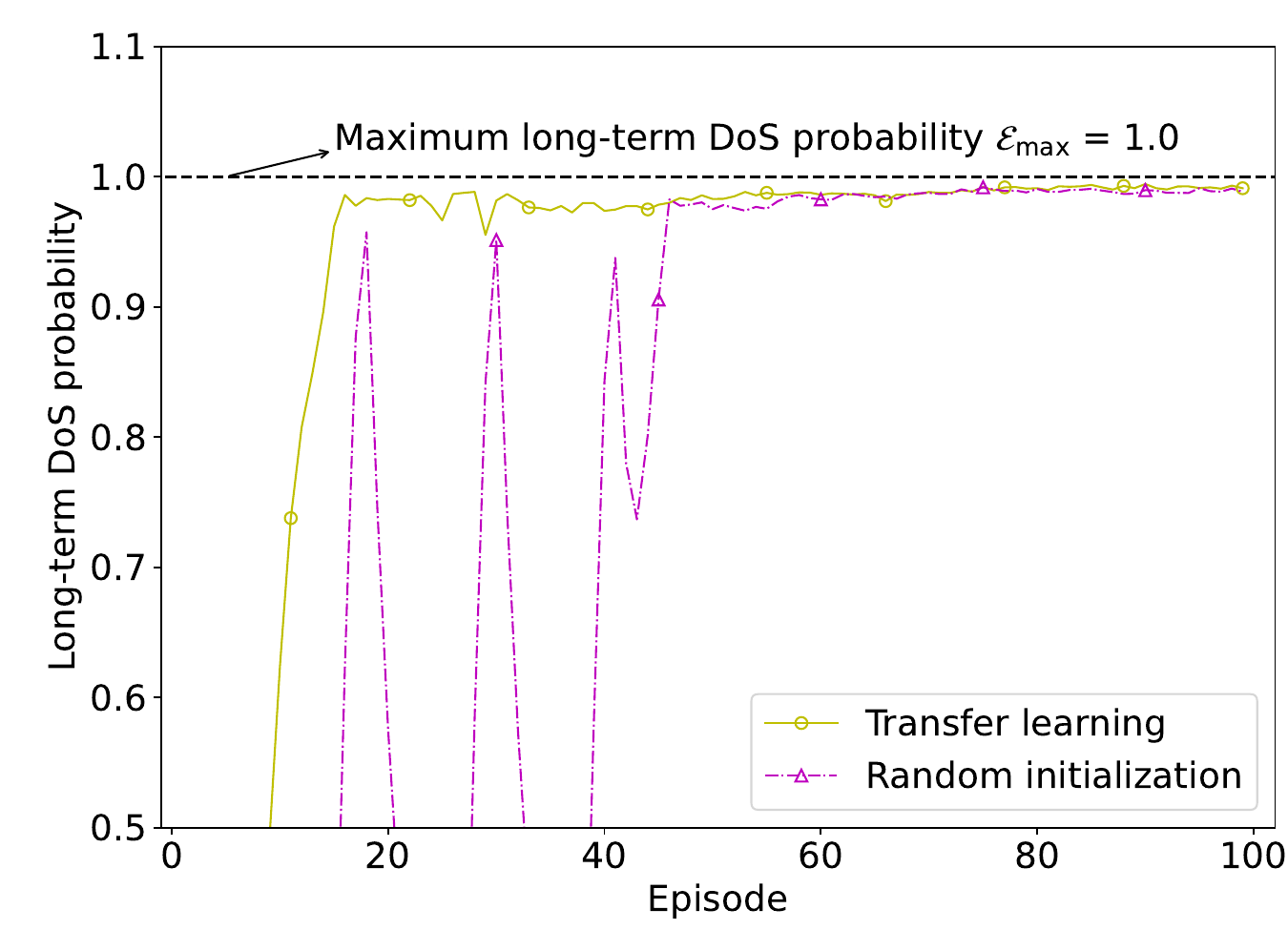}
    \caption{Average DoS probability with random initialization and transfer learning when DoS probability constraint in the long-term equals one.}
    \label{fig_transfer_efficiency_cost}
\end{figure}
Fig.~\ref{fig_transfer_efficiency_cost} plots the training results {of transfer learning and random initialization when the maximum long-term DoS probability, $\mathcal{E}_{\max}$, equals $1$. For transfer learning, we initialize the neural network with the well-optimized parameters of the neural network when $\mathcal{E}_{\max}=0.4$. For the random initialization benchmark, we train the neural network from scratch.} We can observe from Fig.~\ref{fig_transfer_efficiency_cost} that using transfer learning helps the DRL training to converge at approximately 16 episodes, whilst it takes {approximately 47 episodes for random initialization} to converge to approximately the same value. {This is because when the required constraint on DoS probability changes to a new value, the newly updated optimization problem is still related to the previous scenario. Therefore, some of the hidden features that have been well-trained in the previous scenario are still effective to be applied in the new scenario, which further reduces the required training epochs for convergence.}

\begin{figure}[t]
    \centering{\includegraphics[height=6.5cm, width=9cm]{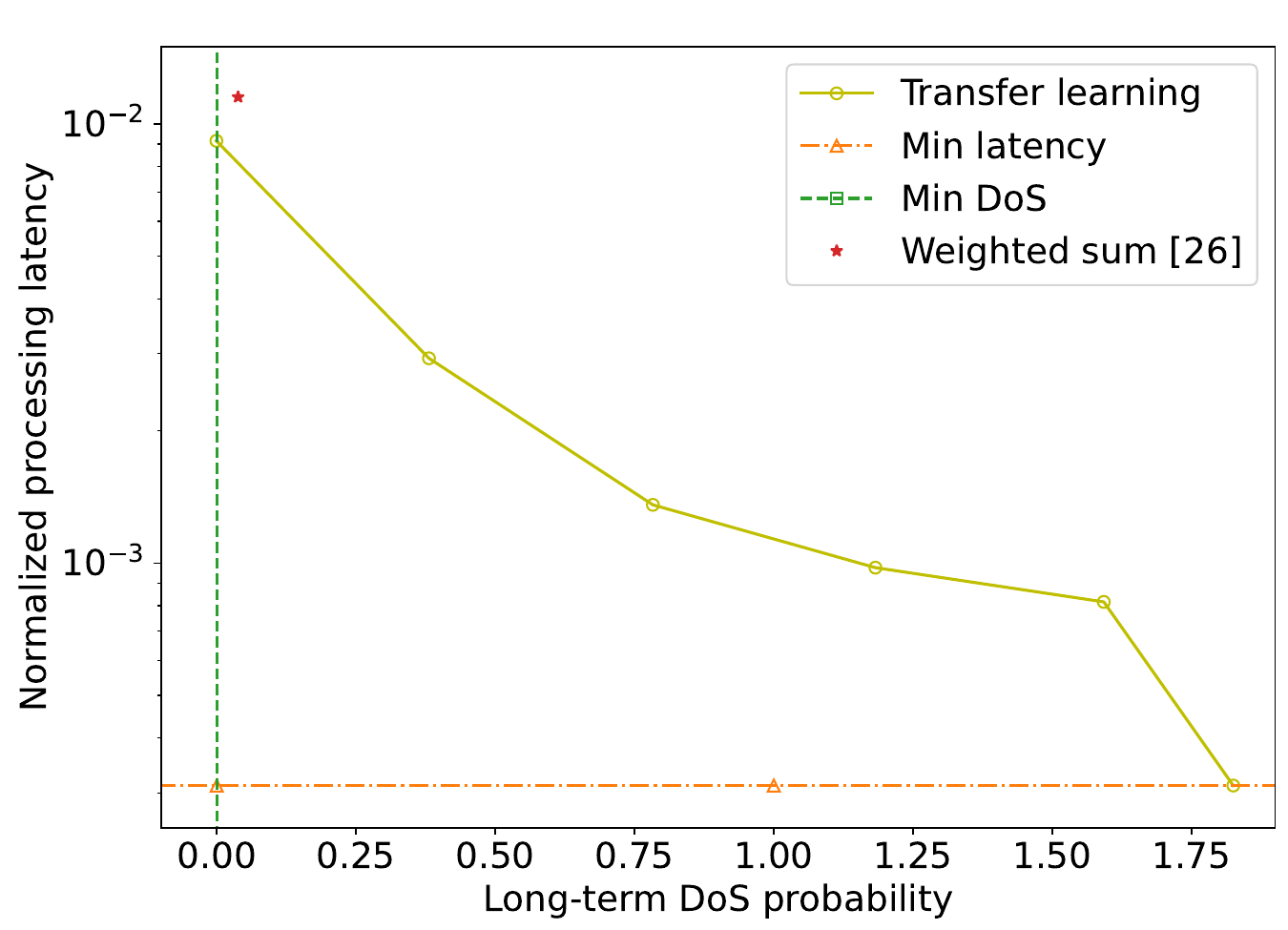}} 
    \caption{{The trade-offs between the processing latency and the long-term DoS probability.}}
    \label{fig_trade_off}
\end{figure}
Fig.~\ref{fig_trade_off} presents the trade-off between the processing latency and the long-term DoS probability achieved through transfer learning. We include a benchmark obtained from a prior study~\cite{RichardYu_MDP_SNRstate}, where the authors applied unconstrained DRL to maximize the weighted sum of the blockchain throughput and the reciprocal of MEC delay, with equal weighting coefficients of $0.5$. It is important to note that the weighted sum benchmark relies on manually selected weighting coefficients and lacks the inherent capability to handle the dynamic constraint. In contrast, our constrained DRL can dynamically update the processing latency in response to changes in the DoS probability constraint. Interestingly, the trade-off figure achieved by transfer learning exhibits some non-smooth behavior. This phenomenon highlights that the converged values achieved by a DRL algorithm are significantly affected by the initialization values.

\section{Conclusion}~\label{sec_conclusion} 
We developed a blockchain-secured deep reinforcement learning (BC-DRL) framework for efficient resource allocation {in dynamic environments}. The BC-DRL {framework} introduced a low-latency reputation-based proof-of-stake (RPoS) blockchain consensus protocol to select trusted base stations (BSs) and resist attacks from both BSs and users. We {formulated the resource optimization problem as a Markov decision process (MDP) that balances} processing latency and BS DoS probability. {To address the challenge of high-dimensional inputs, we designed a} constrained deep reinforcement learning (DRL) algorithm to solve the formulated constrained MDP. 
Numerical experiments {and analysis} showed that the proposed BC-DRL solution requires approximately $2.5$ times less CPU cycles compared with PoW, and can find the optimized resource allocation policy while satisfying the given quality-of-service constraints compared with existing unconstrained DRL-based resource allocation policies. 

\appendices

\renewcommand{\theequation}{A.\arabic{equation}}
\setcounter{equation}{0}
\section{Derivation of Blockchain Processing Latency~\label{Processing_latency_blockchain}}

In the pre-prepare step, the processing latency is expressed as
\begin{equation} 
\begin{split} 
\tau_{\text{bc},i}^\text{g}(t) = 
\dfrac{f_{\text{bc},i}^\text{g}(t)}{a_{i}(t)} 
+\mathop{\max}\limits_{i\in \mathcal{N}_\text{M}(t),j\in \mathcal{N}_\text{V}(t)} \left\{\frac{\ell_\text{b}(t)} {W_{i,j}}\right\},  \text{ (slots)} 
\label{eq: Latency_Generation} 
\end{split} 
\end{equation} 
where the first term is the block miner signature generation time and the second term is the maximum block multicasting time from the miner BS $i\in\mathcal{N}_\text{M}(t)$ to the validator BSs $j\in\mathcal{N}_\text{V}(t)$, where $W_{i,j}$ is the data transmission rate from the $i$-th BS to the $j$-th BS.

In the prepare step, each validator BS adds their signature to the received block from the miner BS. The CPU cycles required for the $j$-th validator in the $t$-th time slot can be calculated by $f_{\text{bc},j}^\text{v}(t)=\mathds{1}\{j \in \mathcal{N}_\text{V}(t)\} \cdot {\kappa}_\text{bc}{\ell_\text{b}(t)}$ (CPU cycles). 
Each validator BS multicasts the block to all the other committee BSs after validating the signature in the block header. Thus, the processing latency is derived as
\begin{equation} 
\begin{split} 
\tau_{\text{bc},i}^\text{v}(t) 
=& 
\mathop{\max}\limits_{j\in \mathcal{N}_\text{V}(t)} \left\{\frac{f_{\text{bc},j}^\text{v}(t)} {a_j(t)}\right\}\\
&+\mathop{\max}\limits_{j\in \mathcal{N}_\text{V}(t),k\in \mathcal{N}_\text{C}(t), k \neq j} \left\{\frac{\ell_\text{b}(t)} {W_{j,k}}\right\}, \quad\text{(slots)} 
\label{eq: Latency_Verification} 
\end{split} 
\end{equation}
where the first term is the maximum signature validating time amongst all validator BSs $j\in \mathcal{N}_\text{V}(t)$, and the second term is the maximum block multicasting time to all committee BSs $k\in \mathcal{N}_\text{C}(t)$.

Finally, in the commit step, the block is multicasted to all other BSs for secure storage in the blockchain. The processing latency is given by
\begin{equation} 
\begin{split} 
\tau_{\text{bc},i}^\text{c}(t) 
= &
\mathop{\max}\limits_{k\in \mathcal{N}_\text{C}(t)} \left\{\frac{f_{\text{bc},k}^\text{c}(t)} {a_{k}(t)}\right\}\\
&+\mathop{\max}\limits_{k\in \mathcal{N}_\text{C}(t),l\in \mathcal{N}_\text{B},l \neq k} \left\{\frac{\ell_\text{b}(t)} {W_{k,l}}\right\}, \quad\text{(slots)} 
\label{eq: Latency_commit} 
\end{split} 
\end{equation}
where the first term is the maximum signature validating time amongst all committee BSs $k\in \mathcal{N}_\text{C}(t)$ and the second term is the maximum multicasting time to all BSs $l\in \mathcal{N}_\text{B}$.

\renewcommand{\theequation}{B.\arabic{equation}}
\setcounter{equation}{0}
\section{Derivation of Service Rate Allocated in One Time Slot}~\label{appendix_singleState}
Based on~\eqref{eq_processing_latency}, the processing latency of the service rates allocated in the $t'$-th time slot is given by
\begin{equation} 
\tau_{i}(t')=\lceil \tau_{\text{bc},i}(t')+\tau_{\text{sp},i}(t') \rceil, \quad \text{(slots)}
\label{eq: reconstuct_tau} 
\end{equation} 
where $\lceil \cdot \rceil$ is the ceiling function. In the $t$-th time slot, the remaining processing latency of the service rates allocated in the $t'$-th time slot can be expressed as
\begin{equation}
\hat{\tau}_{i}(t,t')=(\tau_{i}(t') - (t-t'))^+, \quad \text{(slots)}.
\label{eq: remaining_time}
\end{equation}
In the $t$-th time slot, the service rates been hold equal to the service rates allocated in the $t'$-th time slot, and can be described as
\begin{equation}
\hat{a}_i(t,t') = a_{i}(t') \cdot\mathds{1}\{\hat{\tau}_{i}(t,t') > 0\},  \quad \text{(CPU cycles/slot)}
\end{equation}
where $\mathds{1}\{\cdot\}$ is the indicator function, which equals one when $\hat{\tau}_{i}(t,t') > 0$, and equals zero otherwise. 

The state of the service rates allocated in the $t'$-th time slot is defined as its remaining processing latency and the service rate hold by it, i.e.,
\begin{equation} 
\hat{\boldsymbol{s}}_{i}\left(t,t'\right) = \left[\hat{\tau}_{i}(t,t'),\hat{a}_{i}(t,t')\right]. 
\label{eq: queue_element} 
\end{equation} 
If the BS did not allocate any service rates in the $t'$-th time slot, both $\hat{\tau}_{i}(t,t')$ and ${a}_{i}(t,t')$ are zeros. If the initialized processing latency $\tau_{i}(t')$ is smaller than $T_{\max}$, then $\hat{\tau}_i(t,t')$ and $\hat{a}_i(t,t')$ are also updated to zeros before $t'+T_{\max}$.

\renewcommand{\theequation}{C.\arabic{equation}}
\setcounter{equation}{0}
\section{Proof of Equivalence}~\label{appendix_equivalence}
{We prove the equivalence of problem~\eqref{eq: formulated_optimization} and problem~\eqref{eq: formulated_cmdp} by proving the equivalence of the objective function and the constraint, respectively.}

For the equivalence of the objective function, since our considered stochastic processes are stationary and ergodic, we have $\mathbb{E}[\tau_i(t)] = \mathbb{E}[\tau_i(\hat{t})]$, where $\hat{t} \neq t$. Thus, the objective function can be derived by 
\begin{equation}
\begin{split}
    \max_{\mu(\cdot)} R_{i,\mu}(t)
    =& \max_{\mu(\cdot)} \mathbb{E}_{\mu}\left [\sum\limits_{\hat{t}=t}^{\infty}  {\gamma_r^{\hat{t}-t} r_{i}(t)} \right] \\
    =& \max_{\mu(\cdot)} \mathbb{E}_{\mu} \left[\frac{r_i(t)}{1-\gamma_{r}} \right]\\
    =& \max_{a_i(t)} \mathbb{E}_{\mu} [- \tau_i^\mathrm{a}(t) ]\\
    =& \min_{a_i(t)} \mathbb{E}[\tau_i(t)].
\end{split}
\end{equation}

For the equivalence of the constraint, we take eq.~\eqref{eq: long_term_DoS} and $\mathcal{E}_{\max}={\epsilon_{\max}}/(1-\gamma_{c})$ into eq.~\eqref{eq: DoS_constraint}, we have 
\begin{equation}
C_{i,\mu}(t) = \dfrac{\mathbb{E}_{\mu}[c_{i}(t)]}{1-\gamma_{c}}
\leq 
\frac{\epsilon_{\max}}{1-\gamma_{c}}.
\end{equation}
Thus, we have $\mathbb{E}_{\mu}[c_{i}(t)] \leq \epsilon_{\max}$, which is mathematically equivalent to the constraint in problem~\eqref{eq: formulated_optimization}.

{
In conclusion, the objective function and the constraint of problem~\eqref{eq: formulated_optimization} and problem~\eqref{eq: formulated_cmdp} are all equivalent, thus these two problems are equivalent. This completes the proof.} {\hfill $\square$\par}

\renewcommand{\theequation}{D.\arabic{equation}}
\setcounter{equation}{0}
\section{Proof of the Markov Property}~\label{appendix_MarkovProperty}
In the $t$-th time slot, the action of the $i$-th BS is assigned to provide its physical service rate denoted by $a_{i}(t)$. As shown in~\eqref{eq: state_origin}, the state of the $i$-th BS in the $(t+1)$-th time slot is 
\begin{equation} 
\begin{split}
\hat{\boldsymbol{s}}_{i}(t+1) 
=
\left[ 
\begin{array}{l}   
\boldsymbol{s}_{i}(t+1,t+1-T_{\max})\\  
\boldsymbol{s}_{i}(t+1,t+1-T_{\max}+1)\\  
\vdots\\ 
\boldsymbol{s}_{i}(t+1,t+1)\\ 
\end{array} 
\right]^\mathrm{T}, 
\label{eq: Markov_proof} 
\end{split} 
\end{equation} 
where the dimension of the state in~\eqref{eq: Markov_proof} is $T_{\max}$. Since $s_i(t,t')$ beyond $T_{\max}$ are uncorrelated. Thus, the next state, $\hat{s}_i(t+1)$, only depend on the state and action in the $t$-th time slot. Therefore, the Markov property holds. {This completes the proof.} {\hfill $\square$\par}

\begin{IEEEbiography} 
[{\includegraphics[width=1in,height=1.25in,clip,keepaspectratio]{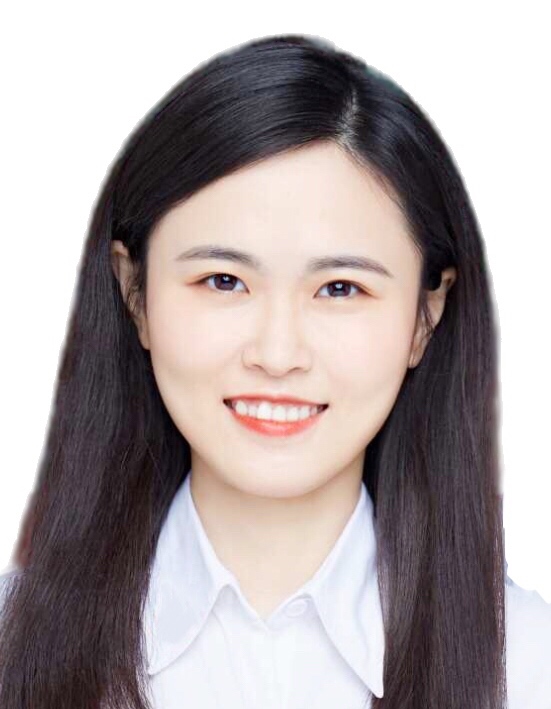}}] 
{Xin Hao} received B.S. and M.E. degrees from the University of Electronic Science and Technology of China (UESTC), respectively. She was a Research Associate at the Microsystem and Terahertz Research Center (MTRC). She is currently pursuing her Ph.D. degree with the School of Electrical and Computer Engineering, University of Sydney (USYD). She is a recipient of the USYD 2023 Faculty of Engineering Career Advancement Award, the 2023 Postgraduate Research Support Scheme, and the 2020 Faculty of Engineering Research Scholarship. Her research interests include blockchain, deep reinforcement learning, graph neural networks, meta-learning, Internet-of-Things, and physical layer security-related wireless communications. She served as a session chair in IEEE ICC 2023. 
\end{IEEEbiography} 
 
\begin{IEEEbiography}[{\includegraphics[width=1in,height=1.25in,clip,keepaspectratio]{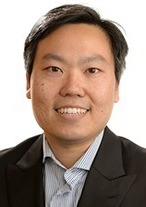}}] 
{Phee~Lep~Yeoh} (S'08-M'12–SM'23)
received the B.E. degree with University Medal and the Ph.D. degree from the University of Sydney, Australia, in 2004 and 2012, respectively. From 2012 to 2016, he was a Lecturer at the University of Melbourne, Australia, and from 2016 to 2023, he was a Senior Lecturer at the University of Sydney, Australia. In 2023, he joined the School of Science, Technology, and Engineering at University of the Sunshine Coast in Queensland, Australia. Dr Yeoh is a recipient of the 2020 University of Sydney Robinson Fellowship, the 2018 Alexander von Humboldt Research Fellowship for Experienced Researchers, and the 2014 Australian Research Council Discovery Early Career Researcher Award. He has also received best paper awards at IEEE PIMRC 2023, IEEE ICC 2014 and IEEE VTC Spring 2013.
\end{IEEEbiography} 

\begin{IEEEbiography}[{\includegraphics[width=1in,height=1.25in,clip,keepaspectratio]{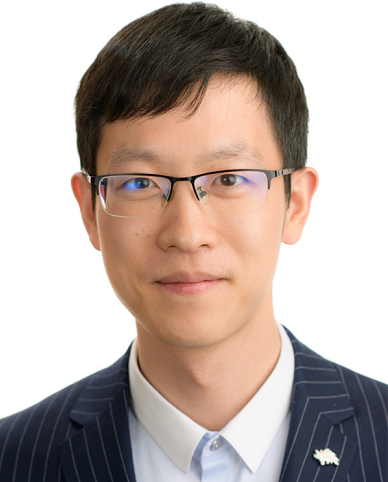}}] 
{Changyang~She} (S'12–M'17–SM'23) received the B.Eng. degree from the Honors College (formerly, School of Advanced Engineering), Beihang University (BUAA), Beijing, China, in 2012, and the Ph.D. degree from the School of Electronics and Information Engineering, BUAA in 2017. From 2017 to 2018, he was a Post-Doctoral Research Fellow with the Singapore University of Technology and Design. Since 2018, he has been a Post-Doctoral Research Associate with the University of Sydney. His research interests lie in the areas of ultrareliable and low-latency communications, tactile Internet, big data for resource allocation in wireless networks, and energy efficient transmission in 5G communication systems.
\end{IEEEbiography} 

\begin{IEEEbiography}[{\includegraphics[width=1in,height=1.25in,clip,keepaspectratio]{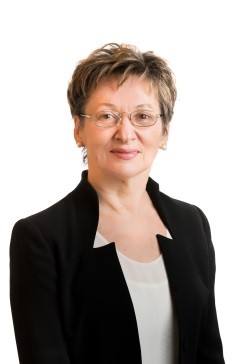}}]{Branka Vucetic} (Life Fellow, IEEE) is currently an ARC Laureate Fellow and Director of the Centre of Excellence for IoT and Telecommunications at the University of Sydney. Her current work is in the areas of wireless networks and Internet of Things. In the area of wireless networks, she works on communication system design for millimetre wave (mmWave) frequency bands. In the area of the Internet of things, she works on providing wireless connectivity for mission critical applications. She is a Fellow of the Australian Academy of Science, the Australian Academy of Technological Sciences and Engineering, and the Engineers Australia.
\end{IEEEbiography}

\begin{IEEEbiography}[{\includegraphics[width=1in,height=1.25in,clip,keepaspectratio]{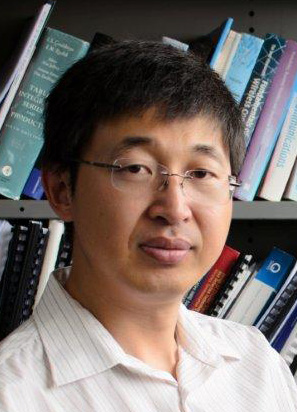}}]{Yonghui~Li} (Fellow, IEEE) received the Ph.D.
degree from Beijing University of Aeronautics and Astronautics, in November 2002. Since 2003, he has been with the Centre of Excellence in Telecommunications, The University of Sydney, Australia. He is now a Professor and Director of Wireless Engineering Laboratory in School of Electrical and Computer Engineering, The University of Sydney. He was the recipient of the Australian Queen Elizabeth II Fellowship in 2008 and the Australian Future Fellowship in 2012. His current research interests include wireless communications, with a particular focus on MIMO, millimeter wave communications, machine to machine communications, coding techniques, and cooperative communications. He holds a number of patents granted and pending in these fields. He was an Editor for IEEE TRANSACTIONS ON COMMUNICATIONS and IEEE TRANSACTIONS ON VEHICULAR TECHNOLOGY. He also served as the Guest Editor for several IEEE journals, such as IEEE JOURNAL ON SELECTED AREAS IN COMMUNICATIONS, IEEE COMMUNICATIONS MAGAZINE, IEEE INTERNET OF THINGS JOURNAL, and IEEE ACCESS. He received the best paper awards from IEEE International Conference on Communications (ICC) 2014, IEEE PIMRC 2017, and IEEE Wireless Days Conferences (WD) 2014.
\end{IEEEbiography}

\end{document}